%% file: root.tex
\documentclass[letterpaper, 10 pt, conference]{ieeeconf}  
\IEEEoverridecommandlockouts 
\usepackage{times}
\usepackage{multicol}
\usepackage[bookmarks=true]{hyperref}
\usepackage{amsmath,amssymb}
\usepackage{graphicx}
\usepackage{stfloats}
\usepackage[dvipsnames]{xcolor}
\usepackage{algorithm}
\usepackage{algpseudocode}
\usepackage{caption}
\usepackage{lettrine}
\usepackage{booktabs}
\usepackage{makecell}
\usepackage{siunitx}
\usepackage{listings} 
\usepackage{xcolor}
\usepackage{wrapfig}

\definecolor{codeblue}{rgb}{0.38039216, 0.61568627, .8       }
\definecolor{codepink}{rgb}{1.        , 0.41568627, 0.83529412}
\definecolor{codegray}{rgb}{0.5,0.5,0.5}
\definecolor{codepurple}{rgb}{0.58,0,0.82}
\definecolor{backcolour}{rgb}{0.98,0.98,0.98}
\lstdefinestyle{mystyle}{
    backgroundcolor=\color{backcolour},   
    commentstyle=\color{codeblue},
    keywordstyle=\color{codepink},
    numberstyle=\tiny\color{codegray},
    stringstyle=\color{codepurple},
    basicstyle=\ttfamily\footnotesize,
    breakatwhitespace=false,         
    breaklines=true,                 
    captionpos=b,                    
    keepspaces=true,                 
    numbers=left,                    
    numbersep=5pt,                  
    showspaces=false,                
    showstringspaces=false,
    showtabs=false,                  
    tabsize=2
}
\lstdefinelanguage{gptprompt}{
  basicstyle=\ttfamily\scriptsize,
  keywords = {generated_example_instruction,generated_example_LTL, given_instruction, stage1_ltl_response, last_response, given_error_clarification, verification_type},
  keywordstyle=\color{red},
  keywords = [2]{isbetween, isabove, isToTheLeft, isbelow, isleftof, isrightof, isnextto, isinfrontof, isbehind},
  keywordstyle= [2]\color{Fuchsia},
  keywords = [3]{Input, Output, Input_instruction, Input_ltl, previous_output, error_clarification, correct_output},
  keywordstyle = [3]\color{blue},
  keywords = [4]{near, pick, release},
  keywordstyle = [4]\color{TealBlue},
}

\lstdefinelanguage{instructions}{
  basicstyle=\ttfamily\scriptsize,
  keywords = {NLMD,NLMC,OKRB,CT,CST},
}

\title{Verifiably Following Complex Robot Instructions with Foundation Models}

\author{Benedict Quartey$^{\dagger*}$, Eric Rosen$^{*}$, Stefanie Tellex, George Konidaris \\ Department of Computer Science, Brown University
\thanks{$^{*}$Equal Contribution}
\thanks{$^{\dagger}$Corresponding Author (\textit{Email}:~\texttt{benedict\_quartey@brown.edu})}
}

\begin{document}

\maketitle

\input{maintext}

\bibliographystyle{IEEEtran}
\bibliography{IEEEabrv,mainref}
\clearpage

\begin{center}
     \large \bf Verifiably Following Complex Robot Instructions with Foundation Models \\  Appendix
\end{center}
\input{appendix}

\end{document}

%% file: maintext.tex
\begin{abstract}
  When instructing robots, users want to flexibly express constraints, refer to arbitrary landmarks, and verify robot behavior, while robots must disambiguate instructions into specifications and ground instruction referents in the real world. To address this problem, we propose Language Instruction grounding for Motion Planning (LIMP), an approach that enables robots to verifiably follow complex, open-ended instructions in real-world environments without prebuilt semantic maps. LIMP constructs a symbolic instruction representation that reveals the robot's alignment with an instructor's intended motives and affords the synthesis of correct-by-construction robot behaviors. We conduct a large-scale evaluation of LIMP on 150 instructions across five real-world environments, demonstrating its versatility and ease of deployment in diverse, unstructured domains. LIMP performs comparably to state-of-the-art baselines on standard open-vocabulary tasks and additionally achieves a 79\% success rate on complex spatiotemporal instructions, significantly outperforming baselines that only reach 38\%. \footnote{See supplementary materials and demo videos at \href{https://robotlimp.github.io/}{\color{blue}{robotlimp.github.io}}}
\end{abstract}

\section{Introduction}
\label{sec:introduction}
\begin{figure}
    \centering
    \includegraphics[width=0.9\columnwidth,trim={0cm 0cm 0cm 0cm},clip]{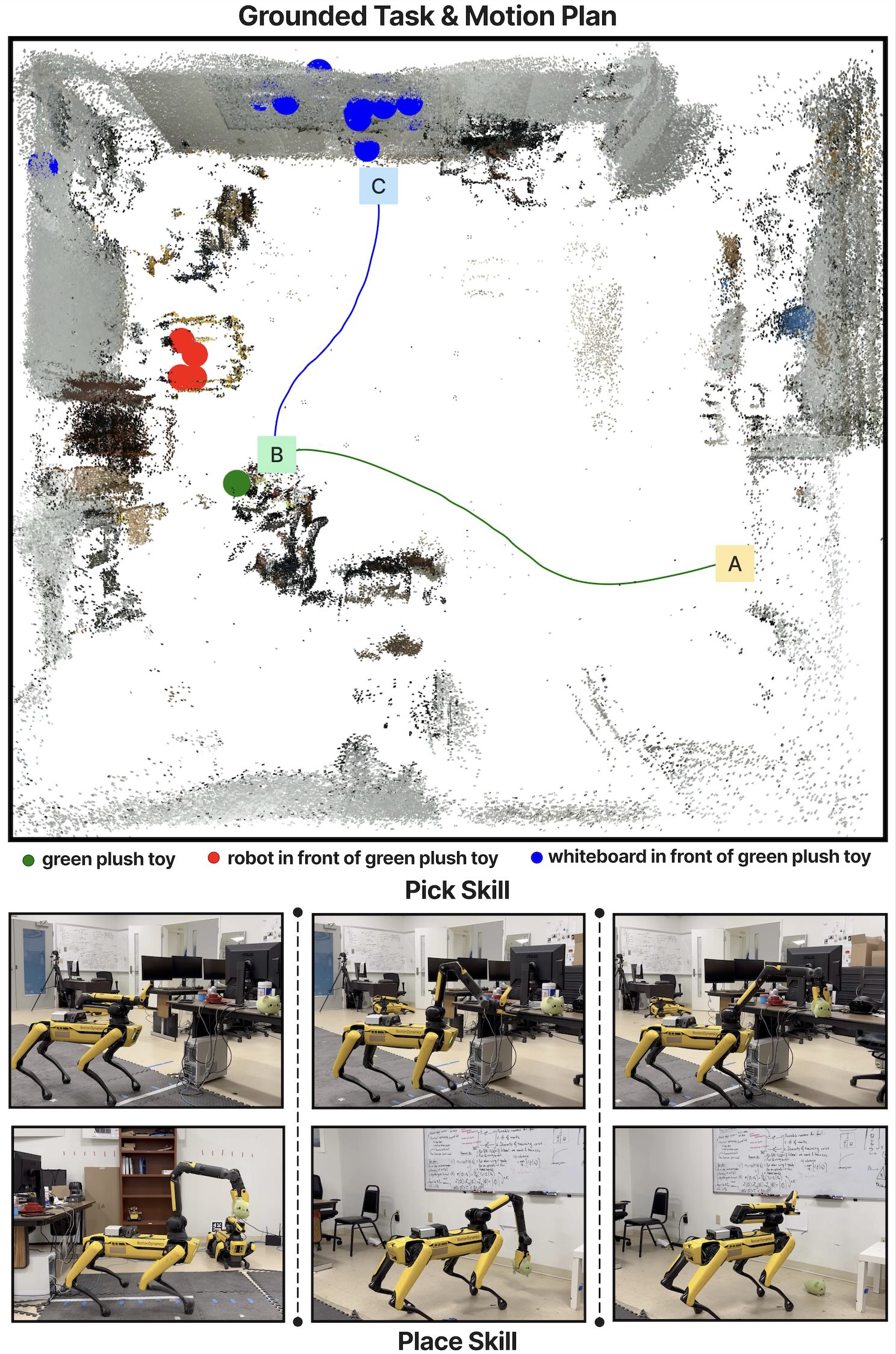}
    \vspace{-0.1cm}
    \caption{Our approach executing the instruction \textit{``Bring the green plush toy to the whiteboard in front of it, watch out for the robot in front of the toy''}. The robot dynamically detects and grounds open-vocabulary referents with spatial constraints to construct an instruction-specific semantic map, then synthesizes a task and motion plan to solve the task. In this example, the robot navigates from its start location (yellow, A), to the green plush toy (green, B), executes a pick skill then navigates to the whiteboard (blue, C), and executes a place skill. Note that the robot has no prior semantic knowledge of the environment.}
    \vspace{-0.6cm}
\label{fig:splash}
\end{figure}

\lettrine{R}{obots} need a rich understanding of natural language to be instructable by non-experts in unstructured environments. People, on the other hand, need to be able to verify that a robot has understood a given instruction and will act appropriately. Achieving these objectives, however, is challenging as natural language instructions often feature ambiguous phrasing, intricate spatiotemporal constraints, and unique referents. To illustrate, consider the instruction shown in Figure~\ref{fig:splash}: \textit{``Bring the green plush toy to the whiteboard in front of it, watch out for the robot in front of the toy''}. Solving such a task requires a robot to ground open-vocabulary referents, follow temporal constraints, and disambiguate objects using spatial descriptions. Foundation models~\cite{hu_toward_2024, firoozi_foundation_2024} offer a path to achieving such complex long-horizon goals; however, existing approaches for robot instruction following have largely focused on navigation~\cite{gadre_cows_2023, shah_lm-nav_2023, huang_visual_2023, huang_audio_2024, liu_grounding_2023}. These methods, broadly classified under object goal navigation~\cite{anderson_evaluation_2018}, enable navigation to instances of an object category but are limited in their ability to localize spatial references and disambiguate object instances based on descriptive language. Other works~\cite{yenamandra_homerobot_2023, chen_open-vocabulary_2023, liu_demonstrating_2024} extend instruction following to mobile manipulation but are limited to tasks with simple temporal constraints expressed in unambiguous language. Moreover, existing efforts typically rely on Large Language Models (LLMs) as complete planners, bypassing intermediate symbolic representations that could provide verification of correctness before execution. Alternative approaches leveraging code-writing LLMs~\cite{huang_visual_2023, huang_audio_2024, liang_code_2023} are susceptible to errors in generated code, which may lead to unsafe robot behaviors. Mapping natural language to specification languages like temporal logic~\cite{emerson_temporal_1990} provides a robust framework for language disambiguation, handling complex temporal constraints, and behavior verification. However, prior works along this line require prebuilt semantic maps with discrete sets of prespecified referents/landmarks from which instructions can be constructed~\cite{liu_grounding_2023, pan_data-efficient_2023, quartey_exploiting_2023}. 

We propose \textit{Language Instruction grounding for Motion Planning (LIMP)}, a method that leverages foundation models and temporal logics to dynamically generate instruction-conditioned semantic maps that enable robots to construct verifiable controllers for following navigation and mobile manipulation instructions with open vocabulary referents and complex spatiotemporal constraints. In a novel environment, LIMP constructs a 3D map via SLAM, then uses LLMs to translate complex natural language instructions into temporal logic specifications with a novel composable syntax for referent disambiguation. Instruction referents are detected and grounded using vision-language models (VLMs) and spatial reasoning. Finally, a task and motion plan is synthesized to guide the robot through the required subgoals, as shown in Figure~\ref{fig:splash}. In summary, we make the following contributions: (\textbf{1}) A modular framework that translates expressive natural language instructions into temporal logic, grounds instruction referents, and executes commands via Task and Motion Planning (TAMP). (\textbf{2}) A spatial grounding method for detecting and localizing open vocabulary objects with spatial constraints in 3D metric maps.
(\textbf{3}) A TAMP algorithm that localizes regions of interest (goal/avoidance zones) and synthesizes constraint-satisfying motion plans for long-horizon tasks. \looseness=-1


\section{Background and Related Works}
\label{sec:related-works}
We briefly highlight the most relevant works in visual scene understanding~\cite{chen_open-vocabulary_2023}, natural language instruction following~\cite{liu_grounding_2023,liu_lang2ltl-2_2024}, and task and motion planning~\cite{rosen_synthesizing_2023}, and provide a comprehensive review in our \href{https://robotlimp.github.io/static/assets/supplementary.pdf}{supplementary materials}. NLMap~\cite{chen_open-vocabulary_2023} grounds open-vocabulary language queries to spatial locations using pre-trained VLMs. While effective for describing individual objects, it cannot handle instructions involving complex constraints between multiple objects due to the lack of object relationship modeling. LIMP addresses this with a novel spatial grounding module that resolves spatial relationships and leverages task and motion planners to satisfy these constraints. Lang2LTL \cite{liu_grounding_2023} is a multi-stage, LLM-based approach that uses entity extraction and replacement to translate language instructions into temporal logic. Its extension \cite{liu_lang2ltl-2_2024} incorporates VLMs and semantic information (via text embeddings) to ground referents. These works require prebuilt semantic maps/databases describing landmarks to ground symbols, whereas our approach dynamically generates landmarks based on open-vocabulary instructions. Action-Oriented Semantic Maps (AOSMs)~\cite{rosen_synthesizing_2023} augment semantic maps with models indicating where robots can perform manipulation skills, integrating with TAMP solvers for mobile manipulation. LIMP similarly provides a TAMP-compatible spatial representation but supports open-vocabulary tasks, whereas AOSMs remain constrained to a fixed set of goals once generated. \looseness=-1
\subsection{Linear Temporal Logic}
\label{sec:background}
LIMP translates natural language instructions into temporal logic specifications for verifiable task and motion planning. While compatible with various specification languages and planning frameworks, we choose Linear Temporal Logic (LTL) \cite{pnueli_temporal_1977} for its proven expressivity in representing complex robot mission requirements \cite{menghi_specification_2021}. LTL defines temporal properties using atomic propositions, logical operators—negation ($\neg$), conjunction ($\land$), disjunction ($\lor$), implication ($\rightarrow$)—and temporal operators: next ($\mathcal{X}$), until ($\mathcal{U}$), globally ($\mathcal{G}$), and finally ($\mathcal{F}$). Despite its expressivity, LTL has been underutilized due to the expert knowledge required to construct specifications, however recent works have seen significant success directly translating natural language into LTL~\cite{liu_grounding_2023, berg_grounding_2020, pan_data-efficient_2023, cosler_nl2spec_2023, fuggitti_nl2ltl_2023, chen_nl2tl_2023}. 
\vspace{1mm}

\noindent\textbf{Behavior Verification}: Expressing instructions as temporal logic specifications allows us to verify the correctness of generated plans a priori. However, instead of explicit verification methods such as model checking, we leverage insights from prior works \cite{vardi_automata-theoretic_1996} and directly use specifications to synthesize plans that are \textit{correct-by-construction} \cite{kress-gazit_temporal-logic-based_2009,colledanchise_synthesis_2017}.

\section{Problem Definition}
\label{sec:problem_definition}
Given a natural language instruction $l$, our goal is to synthesize and sequence navigation and manipulation behaviors to produce a policy that satisfies the temporal and spatial constraints in $l$. Spatial constraints determine task success based on the sequence of robot poses traversed during execution; temporal constraints determine the sequencing of these spatial constraints as a function of task progression. We assume a robot with an RGB-D camera has already navigated a space, capturing images and camera poses. From this data, we build a metric map $m$ (e.g., point cloud, 3D voxel grid) of the environment, defining the space of possible $SE(3)$ poses $P$ and enabling robot localization (i.e., estimating $p_{\text{robot}} \in P$). Unlike previous work leveraging temporal logic~\cite{liu_grounding_2023}, we do not assume access to a semantic map with prespecified object locations or predicates. Instead, we leverage two foundation models: a task-agnostic vision-language model $\sigma$ that, given an image and text, provides bounding boxes or segmentations based on the text; and an auto-regressive large language model $\psi$ that samples likely language tokens based on a history of tokens.
\vspace{1mm}

\noindent\textbf{Navigation}: Navigation is formalized as an object-goal oriented continuous path planning problem, where the goal is to generate paths to a goal pose set $P_{goals} \subset P$ while staying in feasible regions ($P_{feasible} \subset P$) and avoiding infeasible regions ($P_{infeasible} =  P_{feasible}^{C}$). Infeasible regions include environment obstacles as well as dynamically determined semantic regions that violate constraints in the instruction $l$. \looseness=-1
\vspace{1mm}

\noindent\textbf{Manipulation}: We formalize manipulation behaviours as options \cite{sutton_between_1999} parameterized by objects. Consider an object parameter $\theta$ that parameterizes an option $O_{\theta}=(I_{\theta},\pi_{\theta}, \beta_{\theta})$, the initiation set, policy, and termination condition are functions of both the robot pose $P$ and $\theta$. The initiation set $I_{\theta}$ denotes the global reference frame robot positions and object-centric attributes––such as object size––that determine if the option policy $\pi_{\theta}$ can be executed on the object $\theta$. To execute a manipulation skill on an object, an object-goal navigation behavior must first be executed to bring the robot into proximity with the object. We assume access to a library of these manipulation skills and demonstrate our approach on multi-object goal navigation and open-vocabulary mobile pick-and-place~\cite{yenamandra_homerobot_2023,yokoyama_asc_2024}. \looseness=-1


\section{Language Instruction Grounding for Motion Planning}
\begin{figure*}
\centering
\includegraphics[width=1\textwidth]{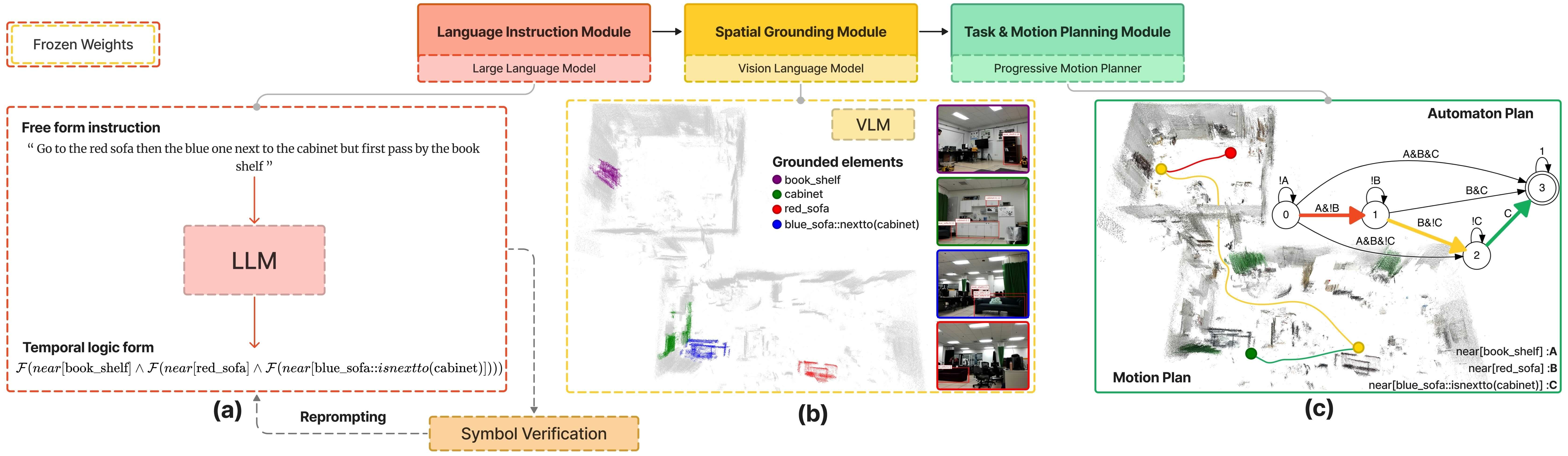}
\captionof{figure}{\textbf{[A]} LIMP translates natural language instructions into temporal logic expressions, where open-vocabulary referents are applied to predicates that correspond to robot skills––note the context-aware resolution of the phrase ``blue one'' to the referent ``blue\_sofa''. \textbf{[B]} Vision-language models detect referents, while spatial reasoning disambiguates referent instances to generate a 3D semantic map that localizes instruction-specific referents. \textbf{[C]} Finally, the temporal logic expression is compiled into a finite-state automaton, which a task and motion planner uses with dynamically-generated task progression semantic maps to progressively identify goals and constraints in the environment, and generate a plan that satisfies the high-level task specification.}
\label{fig:approach}
\vspace{-0.37cm}
\end{figure*}

LIMP interprets expressive natural language instructions to generate instruction-conditioned semantic maps, enabling robots to verifiably solve long-horizon tasks with complex spatiotemporal constraints (Figure~\ref{fig:approach}). We briefly describe our modular approach in this section and present comprehensive implementations details in our \href{https://robotlimp.github.io/static/assets/supplementary.pdf}{supplementary materials}.

\subsection{Language Instruction Module}
\label{sec:language-instruction-section} 
In this module, we leverage a large language model $\psi$ to translate a natural language instruction $l$ into a linear temporal logic specification $\varphi_{l}$ with a novel composable syntax for referent disambiguation. We achieve this through a two-stage in-context learning strategy. The first stage prompts $\psi$ to translate $l$ into a conventional LTL formula $\phi_{l}$ where propositions refer to open-vocabulary objects. The second stage takes $l$ and $\phi_{l}$ as input and prompts $\psi$ to generate a new formula $\varphi_{l}$ with predicate functions corresponding to parameterized robot skills.

We define three predicate functions—\textbf{near}, \textbf{pick}, and \textbf{release}—for the primitive navigation and manipulation skills required for multi-object goal navigation and mobile pick-and-place. Predicate functions in $\varphi_l$ are parameterized by \textit{Composable Referent Descriptors} (CRDs), our novel propositional expressions representing specific referent instances by chaining comparators that encode descriptive spatial information. For example, the instruction \textit{``the yellow cabinet above the fridge that is next to the stove’’} can be represented with the CRD:
\begin{equation}
\text{yellow\_cabinet}::\text{isabove}( \text{fridge}::\text{isnextto}(\text{stove})).
\end{equation}
This specifies that there is a fridge \textit{next to} a stove, and the desired yellow cabinet is \textit{above} that fridge. CRDs are constructed from a set of 3D spatial comparators~\cite{jatavallabhula_conceptfusion_2023} defined in our prompting strategy.

\begin{figure}
\centering
\fbox{\includegraphics[scale=0.19]{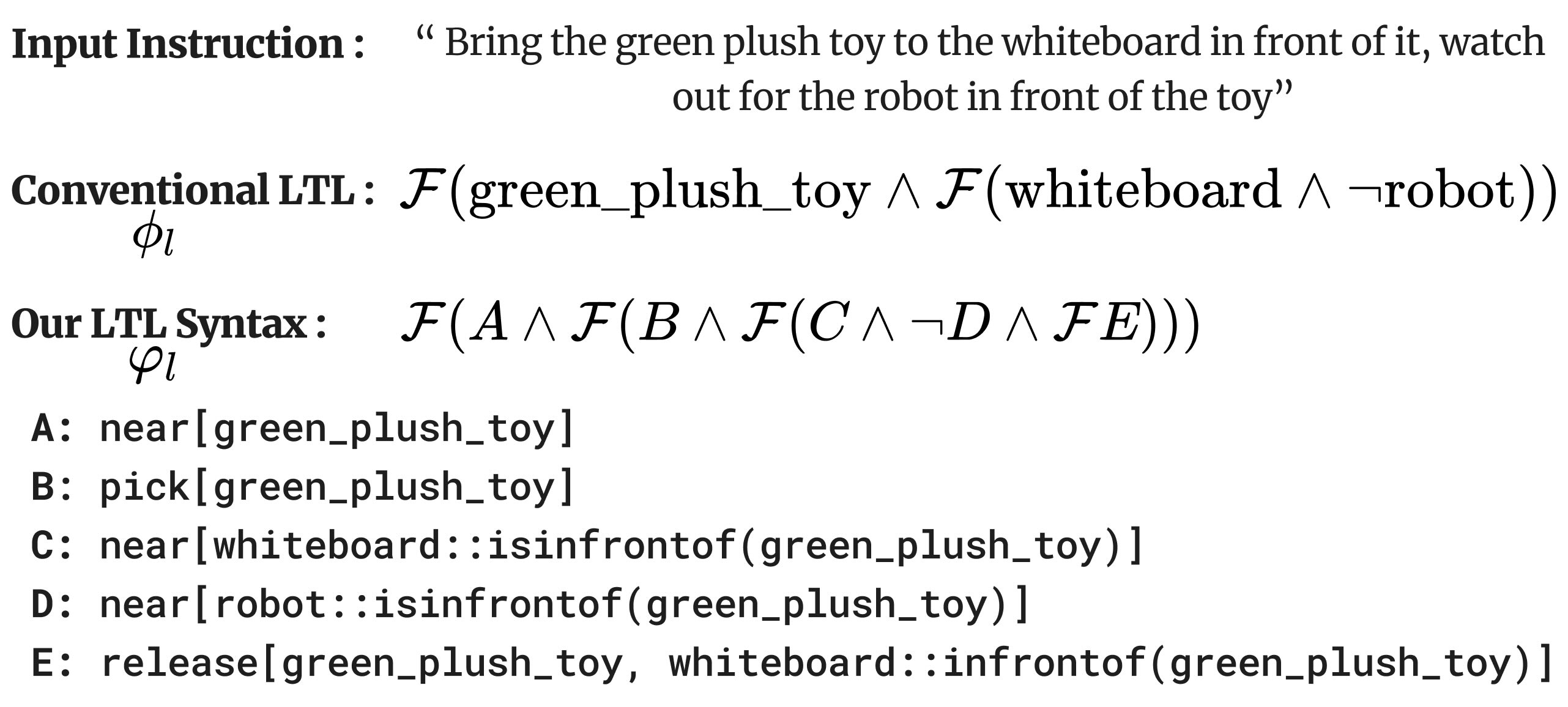}}
\caption{An instruction is first translated into a conventional LTL formula $\phi_l$ that loosely captures the desired temporal occurrence of referent objects, then into our LTL syntax $\varphi_l$ with predicate functions that temporally chain required robot skills parameterized by composable referent descriptors.}
\label{fig:language_module}
\vspace*{-5.5mm}
\end{figure}

Unlike recent works~\cite{yenamandra_homerobot_2023, liu_demonstrating_2024}, our approach does not require specific phrasing or keywords and can handle instructions with arbitrary complexity and ambiguity. The LLM $\psi$ directly samples the entire LTL formula $\varphi_{l}$ with predicate functions parameterized by CRDs using appropriate spatial comparators based on the instruction’s context. Figure~\ref{fig:language_module} illustrates the result of our two-stage prompting strategy.

\vspace{1mm}

\noindent\textbf{LLM Verification}: Verifying the LTL formula $\varphi_{l}$ sampled from the LLM is crucial as errors in referent extraction and temporal task structure affects instruction following accuracy. Our symbol verification node (Figure~\ref{fig:approach}) leverages LTL properties to provide high-level human-in-the-loop verification of extracted instruction referents and temporal task structure. Recent work~\cite{yang_plug_2024} provides ISO 61508~\cite{international2000functional} safety guarantees in robot task execution by translating safety constraints from natural language to LTL formulas, which are verified by human experts and used to enforce robot behavior. Similarly, we rely on human verification to ensure the translated formula $\varphi_{l}$ is correct. Our symbol verification node implements an interactive dialog system that presents users with the extracted referent CRDs and implied task structure, and reprompts the LLM based on user corrections to obtain new formulas. Unlike prior work~\cite{yang_plug_2024}, we eliminate the need for experts by directly translating the task structure—encoded in the LTL formulas's equivalent automaton—back into English statements via a simple deterministic translation scheme. In our experiments (Tables~\ref{tab:instruction-spatial} and~\ref{tab:temporal-tamp}), we find that even without human verification and reprompting, the initial formulas sampled by our language understanding module impressively encode the correct referents and temporal task structure.
\subsection{Spatial Grounding Module}
\label{sec:spatial-grounding-section}
\begin{figure*}
\centering

\includegraphics[width=1\textwidth]{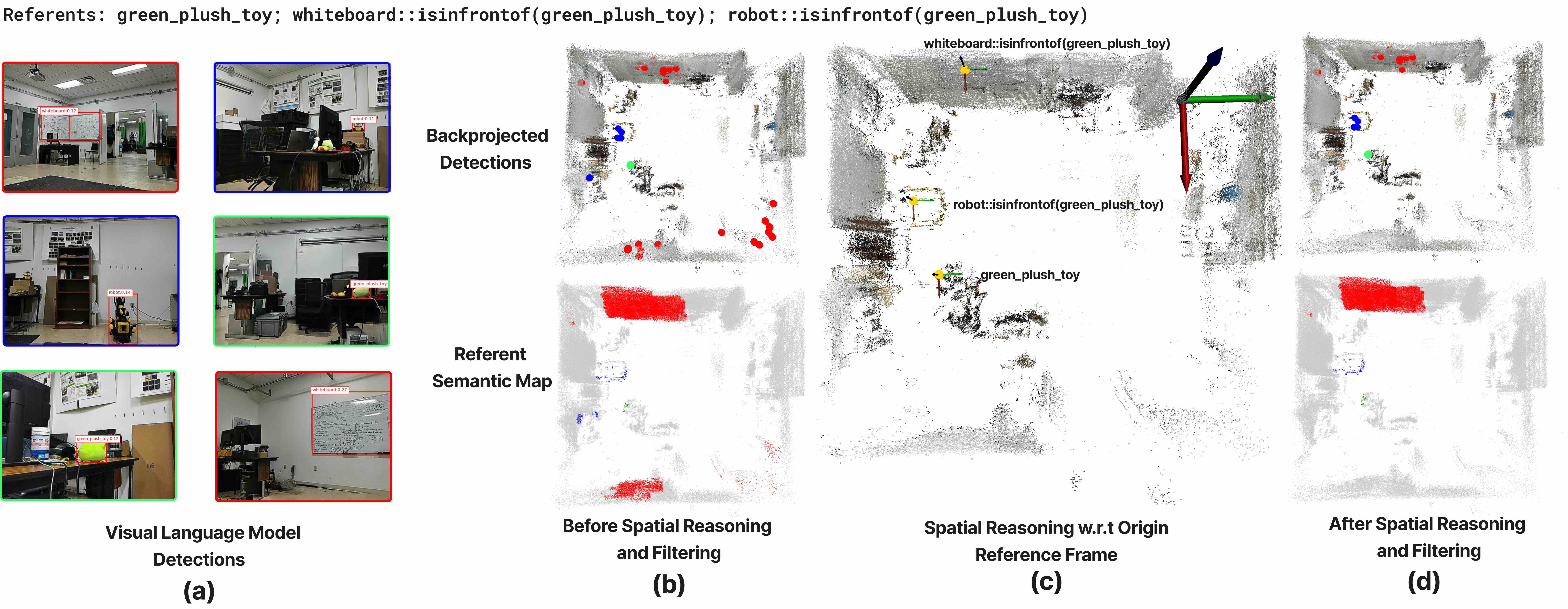}
\vspace{-0.4cm}
\captionof{figure}{\textbf{[A]} Our spatial grounding module leverages a VLM to detect all referent occurrences from prior observations of the environment. \textbf{[B]} An initial semantic map with all detected referent instances is generated by backprojecting pixels in segmented referent masks unto the 3D map. \textbf{[C]} Each referent’s spatial comparators is resolved with respect to the origin coordinate frame of reference. [D] Failing instances are filtered out to obtain a Referent Semantic Map (RSM) that localizes the exact referent instances described in the instruction.}
\label{fig:spatial}
\vspace{-0.3cm}
\end{figure*}
This module detects and localizes specific object instances referenced in an instruction. From the translated LTL formulas, we extract composable referent descriptors (CRDs) and use vision-language models OWL-ViT~\cite{minderer_simple_2022} and SAM~\cite{kirillov_segment_2023} to detect and segment all referent occurrences from the robot's prior observations of the environment. We backproject pixels in these segmentation masks onto our 3D map, creating an initial semantic map of all instruction object instances. From the example in Figure~\ref{fig:language_module}, occurrences of \textit{green\_plush\_toy}, \textit{whiteboard}, and \textit{robot} are detected, segmented, and backprojected onto the map (Figure~\ref{fig:spatial}[a\&b]).

To obtain the specific object instances described in the instruction, we resolve the 3D spatial comparators in each referent’s CRD––recall that CRDs are propositional expressions and can be evaluated as true or false. We define eight spatial comparators (\textit{isbetween}, \textit{isabove}, \textit{isbelow}, \textit{isleftof}, \textit{isrightof}, \textit{isnextto}, \textit{isinfrontof}, \textit{isbehind}) to reason about spatial relationships based on backprojected 3D positions. Since all backprojected positions are relative to an origin coordinate system, our spatial comparators are resolved from the perspective of this origin position as shown in Figure \ref{fig:spatial}[c]. This type of relative frame of reference (FoR) when describing spatial relationships between objects, in contrast to an absolute or intrinsic FoR, is dominant in English \cite{majid_can_2004}, and is a logical choice for our work.

Using the 3D position of each referent’s center mask pixel as its representative position, we resolve a given referent with a spatial description by applying the appropriate spatial comparator to all detected pairs of the desired referent and comparison landmark objects. This filtering process yields a Referent Semantic Map (RSM) that localizes specific object instances described in the instruction as shown in Figure \ref{fig:spatial}[d].

\vspace{1mm}

\noindent\textbf{VLM Verification}: Potential misclassifications from object detector VLMs is the main source of error in this module. We do not address interactively correcting VLM misclassifications as that is out of the scope of this work, but we provide 3D visualization tools that enable users to visually inspect and verify that constructed referent semantic maps correctly localize referents.

\subsection{Task and Motion Planning Module}
\label{sec:task-motion-section}
\begin{figure*}
\centering
\includegraphics[width=1\textwidth]{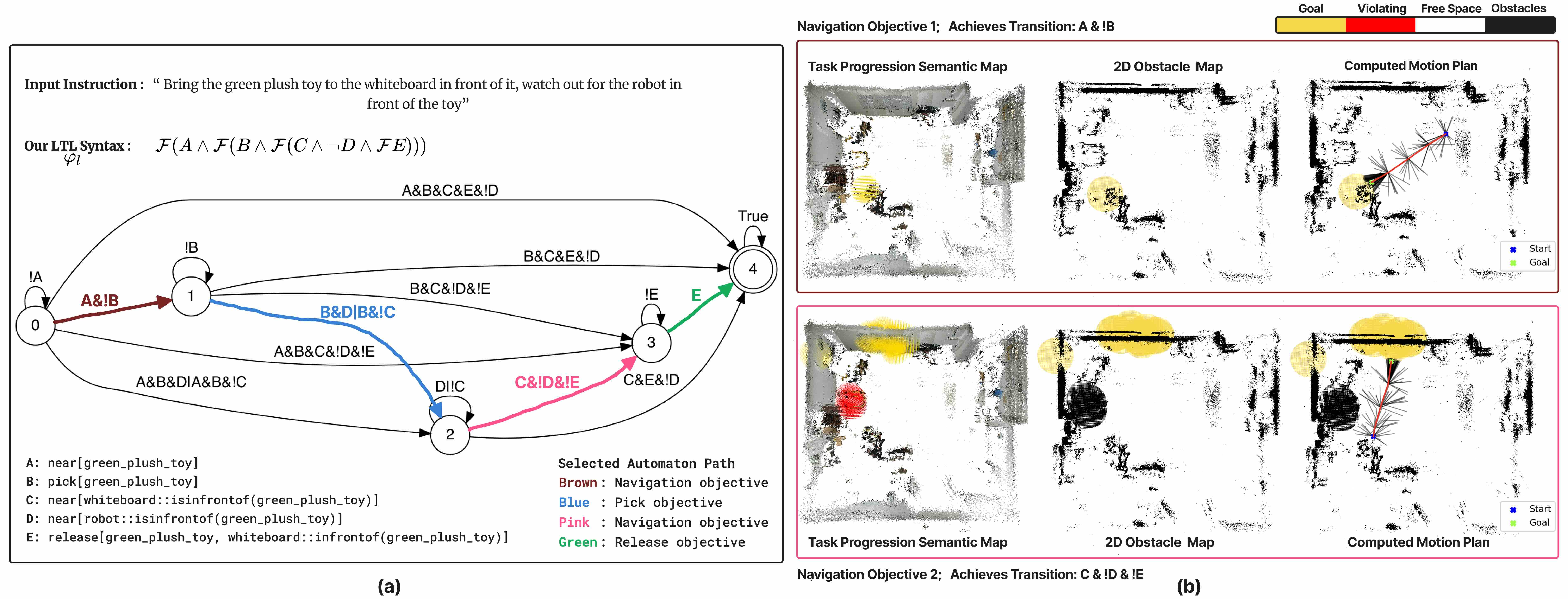}
\vspace{-0.4cm}
\captionof{figure}{\textbf{[A]} A given instruction translated into our LTL syntax $\varphi_l$ can be compiled into an equivalent finite-state automaton that captures the temporal constraints of the task. A path through this automaton is selected with a strategy that incrementally picks the next progression state from the initial state to the accepting state. The robot then executes the manipulation options and navigation behaviors dictated by this high-level task plan. \textbf{[B]} To execute navigation objectives our approach generates a task progression semantic map (TPSM) that augments the environment with state transition constraints, localizing goal (yellow) and avoidance (red) regions. Generated TPSMs are converted into 2D obstacle maps for constraint-aware continuous path planning.}
\label{fig:tamp}
\vspace{-0.5cm}
\end{figure*}

Finally, our TAMP module synthesizes and sequences navigation and manipulation behaviors to produce a plan that satisfies the temporal and spatial constraints expressed in the given instruction.
\vspace{1mm}

\noindent\textbf{Progressive Motion Planner (PMP)}:
Our TAMP algorithm compiles the LTL formula with parameterized robot skills into an equivalent finite-state automaton (Figure~\ref{fig:tamp}[a]) to generate a verifiably correct task and motion plan. A path from the initial to the accepting state in this automaton is a high-level task plan that interleaves navigation and manipulation objectives required to satisfy the instruction. We select such a path with a simple strategy that incrementally selects the next progression state until the accepting state is reached, ensuring the plan obeys all temporal subgoal objectives.  As shown in Figure~\ref{fig:tamp}[a], automaton states are connected by transition edges representing the logical expressions required for transitions. For each transition, our algorithm executes the necessary low-level behaviors: for manipulation subgoals, it executes the appropriate parameterized skill; for navigation subgoals, it dynamically generates Task Progression Semantic Maps (TPSMs) to localize goal and constraint regions and performs continuous path planning using the Fast Marching Tree algorithm (FMT$^*$)~\cite{janson_fast_2015}.

\vspace{1mm}

\noindent\textbf{Task Progression Semantic Maps (TPSM)}: A TPSM augments a 3D scene with navigation constraints specified by logical state transition expressions, enabling goal-directed, constraint-aware navigation. Regions of interest in a TPSM are defined using a nearness threshold specifying proximity to an object. This threshold can be set globally or included in the language instruction module’s prompting strategy, allowing an LLM to infer its value based on the instruction. Like our spatial grounding module, TPSMs are agnostic to temporal logic representations and can be used with various planning approaches for semantic constraint-aware motion planning. We primarily evaluate our approach on a ground mobile robot, hence we transform 3D TPSMs into 2D geometric obstacle maps, where constraint regions are treated as obstacles (Figure~\ref{fig:tamp}[b]). However, our approach is robot-agnostic and supports direct planning in 3D TPSMs for appropriate embodiments like drones. \looseness=-1


\section{Evaluation} 
\label{sec:experiments}
Our evaluations test the hypothesis that translating natural language instructions into LTL expressions and dynamically generating semantic maps enables robots to accurately interpret and execute instructions in large-scale environments without prior training. We focus on three key questions: (1) Can our language instruction module interpret complex, ambiguous instructions? (2) Can our spatial grounding module resolve specific object instances described in instructions? (3) Can our TAMP algorithm generate constraint-satisfying plans?

To answer these questions, we compare LIMP with two baselines: an LLM task planner (NLMap-Saycan~\cite{chen_open-vocabulary_2023}) and an LLM code-writing planner (Code-as-Policies~\cite{liang_code_2023}), representing state-of-the-art approaches for language-conditioned, open-ended robot task execution. Both baselines use the same LLM (GPT-4-0613), prompting structure, and in-context learning examples as our language instruction module. In Code-as-Policies, in-context examples are converted into language model-generated program (LMP) snippets~\cite{liang_code_2023}. To ensure competitive performance, we integrate our CRD syntax, spatial grounding module, and low-level robot control into these baselines, allowing them to query object positions, use our FMT$^*$ path planner, and execute manipulation skills.

We also evaluate ablations of our two-stage language instruction module due to its importance in instruction following. In our full approach, the first stage prompts an LLM to generate a conventional LTL formula $\phi_l$ from instruction $l$ by dynamically selecting relevant in-context examples from a standard dataset \cite{pan_data-efficient_2023} based on cosine similarity. Our first ablation selects in-context examples randomly; and the second ablation skips this stage entirely, directly sampling our LTL syntax with parameterized robot skills $\varphi_l$ from $l$.

We conduct a large-scale evaluation across five real-world environments on a diverse task set of 150 instructions from multiple prior works~\cite{chen_open-vocabulary_2023, liu_demonstrating_2024, liu_grounding_2023}. This task set consists of 24 tasks with fine-grained object descriptions (NLMD), 25 tasks with complex language (NLMC), 25 tasks with simple structured phrasing (OKRB), 37 tasks with complex temporal structures (CT) and 39 tasks with descriptive spatial constraints and temporal structures (CST). Below are examples from each task category illustrating the variety in complexity:
\begin{lstlisting}[language=instructions], 
NLMD: Put the brown multigrain chip bag in the woven basket
NLMC: I like fruits, can you put something I would like on the yellow sofa for me
OKRB: Move the soda can to the box
CT: Visit the purple door elevator, then go to the front desk and then go to the kitchen table, in addition you can never go to the elevator once you have seen the front desk
CST: I have a white cabinet, a green toy, a bookshelf and a red chair around here somewhere. Take the second item I mentioned from between the first item and the third. Bring it the cabinet but avoid the last item at all costs.
\end{lstlisting}


To evaluate instruction understanding, we introduce performance metrics: \textbf{referent resolution accuracy}, \textbf{avoidance constraint resolution accuracy}, and \textbf{spatial relationship resolution accuracy}. These metrics utilize the word error rate (WER), widely used in speech recognition to quantify the difference between a reference and a hypothesis transcription by computing the minimal number of substitutions, deletions, and insertions needed to transform the hypothesis into the reference. WER is calculated as $WER = \frac{S + D + I}{N}$, where  $S$ is substitutions, $D$ deletions,  $I$ insertions, and  $N$ is the total number of words in the reference.  In our work:
\begin{itemize}
\item \textbf{Referent resolution accuracy} compares extracted referents in the generated LTL formula to ground truth referents.
\item \textbf{Avoidance constraint resolution accuracy} compares referents to avoid in the LTL formula (denoted by the unary negation operator) against ground truth avoidance referents.
\item \textbf{Spatial relationship resolution accuracy} compares generated Composable Referent Descriptors (CRDs) in the LTL specification with ground truth CRD expressions.
\end{itemize}

We also define \textbf{temporal alignment accuracy} and \textbf{planning success rate}. A plan is temporally aligned if the sequence of subgoals matches the instructor’s intention, and successful if it satisfies all spatial and temporal constraints specified in the instruction. Achieving a high plan success rate is challenging, requiring accurate referent and avoidance constraint resolution, spatial grounding, and temporal alignment. We report average word error rates for each baseline in Table~\ref{tab:instruction-spatial} and the percentage of successful and temporally accurate plans in Table~\ref{tab:temporal-tamp}. \looseness=-1
\begin{table*}[t]
\centering
\caption{Performance comparison of one-shot instruction understanding and spatial resolution.}
\label{tab:instruction-spatial}
\resizebox{\textwidth}{!}{
\begin{tabular}{@{}lccc@{}}
\toprule
\textbf{Approach} & 
\makecell{Referent Resolution \\ Accuracy \\ (Average WER)  $\downarrow$} & 
\makecell{Avoidance Constraint Resolution \\  Accuracy \\ (Average WER)  $\downarrow$} & 
\makecell{Spatial Relationship Resolution \\  Accuracy \\ (Average WER)  $\downarrow$} \\
\midrule
NLMap-Saycan & 0.09 & 0.12 & 0.05 \\
Code-as-Policies & 0.22 & 0.24 & 0.06 \\
Limp Single Stage Prompting & 0.09 & 0.11 & 0.05 \\
Limp Two Stage Prompting [Random Embedding] & 0.08 & 0.11 & 0.03 \\
\textbf{Limp Two Stage Prompting [Similar Embedding]} &  \textbf{0.07} &  \textbf{0.04} &  \textbf{0.03} \\
\bottomrule
\end{tabular}
}
\vspace{-0.3cm}
\end{table*}

\begin{table*}[t]
\centering
\caption{Performance comparison of one-shot temporal alignment and plan success rate.}
\label{tab:temporal-tamp}
\resizebox{\textwidth}{!}{
\begin{tabular}{@{}lcccccccccc@{}}
\toprule
\textbf{Approach} & 
\multicolumn{5}{c}{\makecell{Temporal Alignment Accuracy \\ (\% of instructions) $\uparrow$}} & 
\multicolumn{5}{c}{\makecell{Planning Success Rate \\ (\% of instructions) $\uparrow$}} \\
\cmidrule(lr){2-6} \cmidrule(lr){7-10}
& NLMD & NLMC & OKRB & CT & CST & NLMD & NLMC & OKRB & CT & CST \\  
\midrule
NLMap-Saycan & 88\% & \textbf{96\%} & 100\% & 32\% & 41\% & 75\% & \textbf{96\%} & 100\% & 35\% & 38\%\\
Code-as-Policies & 58\% & 68\% & 100\% & 35\% & 38\% & 46\% & 68\% &100\% & 38\% & 38\%\\
Limp Single Stage Prompting & 79\% & 64\% & 100\% & 68\% & 74\% & 63\% & 60\% & 100\% & 62\% & 62\%\\
Limp Two Stage Prompting [Random Embedding] & 83\% & 68\% & 100\% & 76\% & 85\% & 79\% & 68\% & 100\% & 57\% & 72\%\\
\textbf{Limp Two Stage Prompting [Similar Embedding]} & \textbf{88\%} & 80\% & \textbf{100\%} & \textbf{76\%} & \textbf{92\%} & \textbf{79\%} & 76\% & \textbf{100\%}  & \textbf{65\%} & \textbf{79\%} \\
\bottomrule
\end{tabular}
}
\vspace{-0.3cm}
\end{table*}

\section{Discussion} 
\label{sec:discussion}
Beyond the verification benefits of symbolic planning, LIMP outperforms baselines in most task sets, notably in complex temporal planning and constraint avoidance. While NLMap-Saycan and Code-as-Policies effectively generate sequential subgoals, they struggle with strict temporal constraints—for example, avoiding a specific referent while approaching another. Our approach ensures each robot step adheres to constraints while achieving subgoals, explaining LIMP’s superior performance on CT and CST tasks. As shown in Table~\ref{tab:temporal-tamp}, LIMP underperforms only against NLMap-Saycan in the NLMC task category. This task set, introduced in the same paper as the baseline \cite{chen_open-vocabulary_2023} (which outperforms LIMP), includes instructions with implicit details such as: “I like fruits, can you put something I would like on the yellow sofa for me.” NLMap-Saycan is better suited to infer and generate plans with possible fruit options, whereas our few-shot LTL translation process is not designed for this.\looseness=-1

\section{Limitations and Conclusion} 
\label{sec:conclusion}
Although LIMP is capable of interpreting non-finite instructions into LTL formulas, our planner is currently limited to processing co-safe formulas, which handle only finite sequences. The accuracy of spatial grounding relies on the performance of vision-language models (VLMs) for object recognition meaning any shortcomings in these systems can negatively impact results. Additionally, LIMP assumes a static environment between mapping and execution, making it not responsive to dynamic changes—an area we aim to address with future work on editable scene representations. Our Progressive Motion Planning algorithm is complete but does not guarantee optimality; however, our framework can be used with existing TAMP planners to enhance efficiency. \looseness=-1

Foundation models hold significant promise for advancing the next generation of autonomous robots. Our results suggest that combining these models—LLMs for language and VLMs for vision—with established methods for safety, explainability, and verifiable behavior synthesis can lead to more reliable and capable robotic systems.\looseness=-1


\section*{Acknowledgement}
This work was supported by the Office of Naval Research (ONR) under REPRISM MURI N000142412603 and ONR \#N00014-22-1-2592, as well as the National Science Foundation (NSF) via grant \#1955361. Partial funding was also provided by The Robotics and AI Institute. \looseness=-1

%% file: appendix.tex
\setcounter{section}{0}
\setcounter{equation}{0}
\setcounter{figure}{0}
\setcounter{table}{0}

\renewcommand{\thesection}{A\arabic{section}}
\renewcommand{\thefigure}{A.\arabic{figure}}
\renewcommand{\thetable}{A.\arabic{table}}
\renewcommand{\theequation}{A.\arabic{equation}}

\section{\textbf{Appendix Summary}}
\label{sec:appendix}
These sections presents additional details on our approach for leveraging foundation models and temporal logics to verifiably follow expressive natural language instructions with complex spatiotemporal constraints without prebuilt semantic maps. We encourage readers to visit our website \href{https://robotlimp.github.io/}{\color{blue}{robotlimp.github.io}} for project summary and demonstration videos. \looseness=-1

\section{Extended Related Works}
\label{sec:extended-review}
\subsection{Foundation Models in Robotics}
Grounding language referents to entities and actions in the world \cite{tellex_robots_2020,blukis_few-shot_2021,patel_grounding_2020,wang_learning_2021} is challenging in part due to the fact that complex perceptual and behavioral meaning can be constructed from the composition of a wide-range of open-vocabulary components\cite{zheng_spatial_2021,berg_grounding_2020,cosler_nl2spec_2023,wang_lana_2023,blukis_few-shot_2021}. To address this problem, foundation models have recently garnered interest as an approach for generating perceptual representations that are aligned with language \cite{park_visual_2023,shah_lm-nav_2023,chen_open-vocabulary_2023,yu_l3mvn_2023,song_llm-planner_2023,huang_visual_2023}. Because there are an ever-expanding number of ways foundation models are being leveraged for instruction following in robotics (e.g: generating plans \cite{huang_visual_2023}, code \cite{liang_code_2023}, etc.), we focus our review on the most related approaches in two relevant application areas: 1) generating natural language queryable scene representations and 2) generating robot plans for following natural language instruction \cite{liu_grounding_2023,hsiung_generalizing_2022}. \looseness=-1

\vspace{1mm}
\noindent\textbf{Visual scene understanding}:
The most similar approach for visual scene understanding to ours is NLMap \cite{chen_open-vocabulary_2023}, a scene representation that affords grounding open-vocabulary language queries to spatial locations via pre-trained visual language models. Given a sequence of calibrated RGB-D camera images and pre-trained visual-language models, NLMap supports language-queries by 1) segmenting out the most likely objects in the 2D RGB images based on the language queries, and 2) estimating the most likely 3D positions via back-projection of the 2D segmentation masks using the depth data and camera pose. While NLMap is suitable for handling complex descriptions of individual objects (e.g: ``green plush toy``), it is fundamentally unable to handle instructions involving complex constraints between multiple objects since it has no way to account for object-object relationships (e.g: ``the green plush toy that is between the toy box and door``). LIMP handles these more complicated language instructions by using a novel spatial grounding module to easily incorporate a wide-variety of complex spatial relationships between objects. In addition, our scene representation is compatible with both LLM planners as well as TAMP solvers, whereas NLMap is only compatible with LLM planners.  \looseness=-1

While NLMap is the most relevant approach to ours, there are other approaches for visual scene understanding and task planning with foundation models which are worth highlighting. VoxPoser \cite{huang_voxposer_2023} leverages the abilities of LLMs to identify affordances and write code for manipulation tasks, along with VLMs complementary abilities to identify open-vocabulary entities in the environment. SayPlan \cite{rana_sayplan_2023} integrates 3D scene graphs with LLM-based planners to bridge the gap between complex, heirarchial scene representations and scalable task planning with open-ended task specifications. Generalizable Feature Fields (GeFF) \cite{qiu_learning_2024} use an implicit scene representation to support open-world manipulation and navigation via an extension of Neural Radiance Fields (NeRFs) and feature distillation in NeRFs. OK-Robot \cite{liu_demonstrating_2024} adopts a system-first approach to solving structured mobile pick-and-place tasks with foundation models by offering an integrated solution to object detection, mapping, navigation and grasp generation. While these methods are related, none of them have all the features of LIMP: 1) Explicit support for both LLM-based planning and off-the-shelf task and motion planning approaches, 2) Verifiable representations for following complex natural language instructions in mobile manipulation domains that involve object-object relationships, and 3) The ability to dynamically generate task-relevant state abstractions (semantic maps) for individual instructions. \looseness=-1

\vspace{1mm}
\noindent\textbf{Language instruction for robots}:
Our approach to handling complex natural language instructions involves translating the command into a temporal logic expression. This problem framing allows us to leverage state-of-the-art techniques from machine translation, such as instruction-tuned large language models. Most similar to our approach in this regard is \cite{liu_grounding_2023}, which uses a multi-stage LLM-based approach and finetuning to perform entity-extraction and replacement to translate natural language instructions into temporal logic expressions. However, \cite{liu_grounding_2023} relies on a prebuilt semantic map that grounds expression symbols, limiting the scope of instructions it can operate since landmarks are predetermined. Instead, our approach interfaces with a novel scene representation that supports open-vocabulary language and generates the relevant landmarks based on the open-vocabulary instruction. Additionally, the symbols in our temporal logic translation correspond to parameterized task relevant robot skills as opposed to propositions of referent entities extracted from instructions. \looseness=-1

\subsection{Planning Models in Robotics}
\noindent\textbf{Semantic Maps}: Semantic maps \cite{kostavelis_semantic_2015} are a class of scene representations that capture semantic (and typically geometric) information about the world, and can be used in cojunction with planners to generate certain types of complex robot behavior like collision-free navigation with spatial constraints \cite{crespo_semantic_2020,pronobis_semantic_2011}. However, leveraging semantic maps for task planning with mobile manipulators has been challenging since  the modeling information needed may highly depend on the robot's particular skills and embodiment. \cite{rosen_synthesizing_2023} recently proposed Action-Oriented Semantic Maps (AOSMs), which are a class of semantic maps that include additional models of the regions of space where the robot can execute manipulation skills (represented as symbols). \cite{rosen_synthesizing_2023} demonstrated that AOSMs can be used as a state representation that supports TAMP solvers in mobile manipulation domains. Our scene representation is similar to an AOSM since it captures spatial information about semantic regions of interest, and is compatible with TAMP solvers, but largely differs in that AOSMs require learning via online interaction with the scene. Instead, our approach leverages foundation models and requires no online learning. Also, once an AOSM is generated for a scene, there is only a closed-set of goals that can be planned for, whereas our approach can handle open-vocabulary task specifications. \looseness=-1

\vspace{1mm}
\noindent\textbf{Task and Motion Planning}: Task and Motion planning approaches are hierarchical planning methods that involve high-level task planning (with a discrete state space) \cite{fikes_strips_1971} and low-level motion planning (with a continuous state space) \cite{garrett_integrated_2021}. The low-level motion planning problem involves generating paths to goal sets through continuous spaces (e.g: configuration space, cartesian space) with constraints on infeasible regions. When the constraints and dynamics can change, it is referred to as multi-modal motion planning, which naturally induces a high-level planning problem that involves choosing which sequence of modes to plan through, and a low-level planning problem that involves moving through a particular mode. Finding high-level plan skeletons and satisfying low-level assignment values for parameters to achieve goals is a challenging bi-level planning problem\cite{garrett_integrated_2021}. LIMP contains sufficient information to produce a problem and domain description augmented with geometric information for bi-level TAMP solvers like \cite{garrett_pddlstream_2020,holladay_planning_2021}. \looseness=-1

\section{Language Instruction Module}
\label{sec:instruction-module-details}
We implement a two-stage prompting strategy in our language instruction module to translate natural language instructions into LTL specifications. The first stage translates a given instruction into a conventional LTL formula, where propositions refer to open-vocabulary objects. For any given instruction, we dynamically generate $K$ in-context translation examples from a standard dataset \cite{pan_data-efficient_2023} of natural language and LTL pairs, based on cosine similarity with the given instruction. Here is the exact text prompt used:
\begin{lstlisting}[language=gptprompt, caption=Base prompt used to obtain a conventional LTL formula from a natural language query]
You are a LLM that understands operators involved with Linear Temporal Logic (LTL), such as F, G, U, &, |, ~ , etc. Your goal is translate language input to LTL output.
Input:<generated_example_instruction>
Output:<generated_example_LTL>
...
Input:<given_instruction>
Output:
\end{lstlisting}

The second stage takes the given instruction and the LTL response from the first stage as input to generate a new LTL formula with predicate functions that correspond to parameterized robot skills. Skill parameters are instruction referent objects expressed in our novel Composable Referent Descriptor (CRD) syntax. CRDs enable referent disambiguation by chaining comparators that encode descriptive spatial information. We define eight spatial comparators and provide their descriptions as part of the second stage prompt. We find that LLMs conditioned on this information and a few examples are able translate arbitrarily complex instructions with appropriate comparator choices. Here is the exact prompt used: \looseness=-1
\begin{lstlisting}[language=gptprompt, caption= Second stage prompt to output our LTL syntax with CRD parameterized robot skills]
You are an LLM for robot planning that understands operators involved with Linear Temporal Logic (LTL), such as F, G, U, &, |, ~ , etc. You have a finite set of robot predicates and spatial predicates, given a language instruction and an LTL formula that represents the given instruction, your goal is to translate the ltl formula into one that uses appropriate composition of robot and spatial predicates in place of propositions with relevant details from original instruction as arguments.
Robot predicate set (near,pick,release). 
Usage:
near[referent_1]:returns true if the desired spatial relationship is for robot to be near referent_1.
pick[referent_1]:can only execute picking skill on referent_1 and return True when near[referent_1].
release[referent_1,referent_2]:can only execute release skill on referent_1 and return True when near[referent_2].
Spatial predicate set (isbetween,isabove,isbelow,isleftof,isrightof,isnextto,isinfrontof,isbehind). 
Usage:
referent_1::isbetween(referent_2,referent_3):returns true if referent_1 is between referent_2 and referent_3.
referent_1::isabove(referent_2):returns True if referent_1 is above referent_2.
referent_1::isbelow(referent_2):returns True if referent_1 is below referent_2.
referent_1::isleftof(referent_2):returns True if referent_1 is left of referent_2.
referent_1::isrightof(referent_2):returns True if referent_1 is right of referent_2.
referent_1::isnextto(referent_2):returns True if referent_1 is close to referent_2.
referent_1::isinfrontof(referent_2):returns True if referent_1 is in front of referent_2.
referent_1::isbehind(referent_2):returns True if referent_1 is behind referent_2.
Rules:
Strictly only use the finite set of robot and spatial predicates!
Strictly stick to the usage format!
Compose spatial predicates where necessary!
You are allowed to modify the structure of Input_ltl for the final Output if it does not match the intended Input_instruction!
You should strictly only stick to mentioned objects, however you are allowed to propose and include plausible objects if and only if not mentioned in instruction but required based on context of instruction! 
Pay attention to instructions that require performing certain actions multiple times in generating and sequencing the predicates for the final Output formula!
Example:
Input_instruction: Go to the orange building but before that pass by the coffee shop, then go to the parking sign.
Input_ltl: F (coffee_shop & F (orange_building & F parking_sign ) )
Output: F ( near[coffee_shop] & F ( near[orange_building] & F near[parking_sign] ))
Input_instruction: Go to the blue sofa then the laptop, after that bring me the brown bag between the television and the kettle on the left of the green seat, I am standing by the sink.
Input_ltl: F ( blue_sofa & F ( laptop & F ( brown_bag & F ( sink ) ) ) )
Output: F ( near[blue_sofa] & F ( near[laptop] & F ( near[brown_bag::isbetween(television,kettle::isleftof(green_seat))] & F (pick[brown_bag::isbetween(television,kettle::isleftof(green_seat))] & F ( near[sink] & F ( release[brown_bag,sink] ) ) ) ) ) )
Input_instruction: Hey need you to pass by chair between the sofa and bag, pick up the bag and go to the orange napkin on the right of the sofa.
Input_ltl: F ( chair & F ( bag & F ( orange_napkin ) ) )
Output: F ( near[chair::isbetween(sofa,bag)] & F ( near[bag] & F ( pick[bag] & F ( near[orange_napkin::isrightof(sofa)] ) ) ) )
Input_instruction:  Go to the chair between the green laptop and the yellow box underneath the play toy
Input_ltl: F ( green_laptop & F ( yellow_box & F ( play_toy & F ( chair ) ) ) )
Output: F ( near[chair::isbetween(green_laptop,yellow_box::isbelow(play_toy))] )
Input_instruction: Check the table behind the fridge and bring two beers to the couch one after the other
Input_ltl: F ( check_table & F ( bring_beer1 ) & F ( bring_beer2 ) & F ( couch ) )
Output: F ( near[table::isbehind(fridge)] & F ( pick[beer] & F ( near[couch] & F ( release[beer,couch] & F ( near[table::isbehind(fridge)] & F ( pick[beer] & F ( near[couch] & F ( release[beer,couch] ))))))))
Input_instruction: <given_instruction>
Input_ltl: <stage1_ltl_response>
Output:
\end{lstlisting}

\subsection{Interactive Symbol Verification}
\label{sec:interactive-verification-details}
Verifying sampled LTL formulas is essential, as such we implement an interactive dialog system that presents users with extracted referent composible referent descriptors (CRDs) in sampled formulas as well as the implied task structure––encoded in the sequence of state-machine transition expressions that must hold to progressively solve the task. We translate the task structure into English statements via a simple deterministic strategy that replaces logical connectives and skill predicates from the formula with equivalent English phrases. Users can verify a formula as correct or provide corrective statements which are used to reprompt the LLM to obtain new formulas.  Below is the exact prompt used for regenerating formulas. \looseness=-1
\begin{lstlisting}[language=gptprompt, caption=Corrective reprompting prompt used to obtain new LTL formulas]
There was a mistake with your output LTL formula: Error with <verification_type>! Consider the clarification feedback and regenerate the correct output for the Input_instruction. Make sure to adhere to all rules and instructions in your original prompt!
previous_output:<last_response>
error_clarification: <given_error_clarification>
correct_output:
\end{lstlisting}
To illustrate, the instruction \textit{``Bring the green plush toy to the whiteboard in front of it"} yields the interactive Referent and Task Structure Verification dialog below:
\begin{lstlisting}[language=gptprompt, caption=Interactive referent and task structure verification dialog.]
**************************
Instruction Following
**************************
Input_instruction: "Bring the green plush toy to the whiteboard in front of it" 
Sampled LTL formula: F(A & F(B & F(C & FD)))
    A: near[green_plush_toy]
    B: pick[green_plush_toy]
    C: near[whiteboard::isinfrontof(green_plush_toy)]
    D: release[green_plush_toy, whiteboard::isinfrontof(green_plush_toy)]

***************************
Referent Verification
*************************** 
I extracted this list of relevant objects based on your instruction:
    * whiteboard::isinfrontof(green_plush_toy) 
    * green_plush_toy
Does this match your intention? (y/n)

****************************
Task Structure Verification
****************************
Based on my understanding here is the sequence of subgoal objectives needed to satisfy the task:
Subgoal_1:
    Logical Expression: A&!B
    English translation: I should be near the [green_plush_toy] and not have picked up the [green_plush_toy] 
Subgoal_2:
    Logical Expression: B&!C
    English translation: I should have picked up the [green_plush_toy] and not be near the [whiteboard::isinfrontof(green_plush_toy)] 
Subgoal_3:
    Logical Expression: C&!D
    English translation: I should be near the [whiteboard::isinfrontof(green_plush_toy)] and not have released the [green_plush_toy] at the [whiteboard::isinfrontof(green_plush_toy)]
Subgoal_4:
    Logical Expression: D
    English translation: I should have released the [green_plush_toy] at the [whiteboard::isinfrontof(green_plush_toy)] 
Does this match your intention? (y/n)
\end{lstlisting}

\section{Spatial Grounding Module}
\label{sec:spatial-grounding-details}
The spatial grounding module detects and localizes specific instances of objects referenced in a given instruction by first detecting, segmenting and back-projecting all referent occurances and then filtering based on the descriptive spatial details captured by each referent's composable referent descriptor (CRD). We use the Owl-Vit model \cite{minderer_simple_2022} to detect bounding boxes of open-vocabulary referents and SAM \cite{kirillov_segment_2023} to generate masks from detected bounding boxes. To illustrate referent filtering via spatial information, consider an example scenario where the goal is to resolve the composable referent descriptor below: 
\begin{equation}
\text{whiteboard}::isinfrontof( \text{green\_plush\_toy}).
\end{equation}
Let $W = \{w_1, w_2, \ldots, w_n\}$ and $G = \{g_1, g_2, \ldots, g_m\}$ represent the set of representative 3D positions of detected whiteboards and green\_plush\_toys respectively. The cartesian product of these sets enumerates all possible pairs $(w, g)$ for comparison.
\begin{equation}
    W \times G = \{ w, g) \mid w \in W, g \in G \}
\end{equation}

The `isinfrontof(w, g)' comparator is applied to each pair, yielding a subset \( S \) that contains only those `whiteboards` that satisfy the `isinfronto' condition with at least one `green\_plush\_toy'.
\begin{equation}
    S = \{ w \in W \mid \exists g \in G \text{ such that } \text{isinfrontof}(w, g) \text{ is true} \}
\end{equation}

\subsection{3D Spatial Comparators} 
Our 3D spatial comparators enable Relative Frame of Reference (FoR) spatial reasoning between referents, based on their backprojected 3D positions. Threshold values in the spatial comparators give users the ability to specify the sensitivity or resolution at which spatial relationships are resolved, we keep all threshold values fixed across all experiments. Below is a description of each spatial comparator. \looseness=-1

\begin{lstlisting}[language=gptprompt, caption= Implementation description of 3D spatial comparators]
1. isbetween(referent_1_pos, referent_2_pos, referent_3_pos, threshold): Returns true if referent_1 is within 'threshold' distance from the line segment connecting referent_2 to referent_3, ensuring it lies in the directional path between them without extending beyond.
2. isabove(referent_1_pos, referent_2_pos, threshold): Returns true if the z-coordinate of referent_1 exceeds that of referent_2 by at least 'threshold'.
3. isbelow(referent_1_pos, referent_2_pos, threshold): Returns true if the z-coordinate of referent_1 is less than that of referent_2 by more than 'threshold'.
4. isleftof(referent_1_pos, referent_2_pos, threshold): Returns true if the y-coordinate of referent_1 exceeds that of referent_2 by at least 'threshold', indicating referent_1 is to the left of referent_2.
5. isrightof(referent_1_pos, referent_2_pos, threshold): Returns true if the y-coordinate of referent_1 is less than that of referent_2 by more than 'threshold', indicating referent_1 is to the right of referent_2.
6. isnextto(referent_1_pos, referent_2_pos, threshold): Returns true if the Euclidean distance between referent_1 and referent_2 is less than 'threshold', indicating they are next to each other.
7. isinfrontof(referent_1_pos, referent_2_pos, threshold): Returns true if the x-coordinate of referent_1 is less than that of referent_2 by more than 'threshold', indicating referent_1 is in front of referent_2.
8. isbehind(referent_1_pos, referent_2_pos, threshold): Returns true if the x-coordinate of referent_1 exceeds that of referent_2 by at least 'threshold', indicating referent_1 is behind referent_2.
\end{lstlisting}

\section{Task and Motion Planning Module} 
\label{sec:planning-module-details}
We present pseudocode for our Progressive Motion Planner (Alg.\ref{alg:tamp-algo}) and our algorithm for generating Task Progression Semantic Maps (Alg.\ref{alg:tpsm-algo}). Alg.\ref{alg:tpsm-algo} generates a TPSM $\mathcal{M}_{\text{tpsm}}$ by integrating an environment map ($\mathcal{M}$) and a referent semantic map ($\mathcal{M}_{\text{rsm}}$) given a logical transition expression ($\mathcal{T}$), a desired automaton state ($\mathcal{S'}$), and a nearness threshold ($\theta$). The algorithm first initializes $\mathcal{M}_{\text{tpsm}}$ with a copy of $\mathcal{M}$ and extracts relevant instruction predicates from $\mathcal{T}$. For each predicate (parameterized skill), the algorithm identifies satisfying referent positions in $\mathcal{M}_{\text{rsm}}$, generates a spherical grid of surrounding points within a radius $\theta$, and assesses how these points affect the progression of the task automaton towards $\mathcal{S'}$. These points demarcate regions of interest, and are assigned a value of \textit{1} if they cause the automaton to transition to the desired state, \textit{-1} if they lead to a different automaton state or violate the automaton, and \textit{0} if they do not affect the automaton. The points are then integrated into $\mathcal{M}_{\text{tpsm}}$, yielding a semantic map that identifies goal and constraint violating regions.
\vspace{2mm}
\begin{algorithm}[]
\caption{Progressive Motion Planning Algorithm}

\label{alg:tamp-algo}
\begin{algorithmic}[1]
\Procedure{PMP}{$X_{start}, \varphi, \mathcal{M}, \mathcal{M}_{rsm}, \theta$}
    \Statex \textbf{Input:}
    \Statex \quad \( X_{start} \): Start position in the environment.
    \Statex \quad \( \varphi \): CRD syntax LTL formula specifying task objectives.
    \Statex \quad \( \mathcal{M} \): Environment map.
    \Statex \quad \( \mathcal{M}_{rsm} \): Referent semantic map.
    \Statex \quad \( \theta \): Nearness threshold.

    \Statex \textbf{Output:}
    \Statex \quad \( \Pi \): Generated task and motion plan.

    \State \( \mathcal{A} \leftarrow \text{ConstructAutomaton}(\varphi) \)
    \State \( \text{path} \leftarrow \text{SelectAutomatonPath}(\mathcal{A}) \) \Comment{Task plan}

    \While{$\Pi$.\text{status is active}}
        \While{$\mathcal{A}$.state != path.acceptingState}
        \State \( \mathcal{S}, \mathcal{T}, \mathcal{S'} \leftarrow \mathcal{A}.\text{GetTransition(path.currentStep)} \)
            \State \( \text{objective} \leftarrow \text{NextObjectiveType}(\mathcal{T}) \)

            \If{\( \text{objective} = \text{``skill''} \)}
                \State \( \Pi.\text{UpdateWithSkill}(\mathcal{T}) \)
                \State $\mathcal{A}$.UpdateAutomatonState($\mathcal{S'}$)
            \ElsIf{\( \text{objective} = \text{``navigation''} \)}
                \State \( \mathcal{M}_{params} \leftarrow \mathcal{M}, \mathcal{M}_{rsm}, \mathcal{T}, \mathcal{A}, \mathcal{S'}, \theta \)
                \State \( \mathcal{M}_{tpsm} \leftarrow \Call{GenerateTPSM}{\mathcal{M}_{params}} \)
                \State \( \mathcal{O} \leftarrow \text{GenerateObstacleMap}(\mathcal{M}_{tpsm}) \)
                \State \( \text{plan} \leftarrow \text{FMT}^*(X_{start}, \mathcal{O}) \) \Comment{Path plan}
                \If{\( \text{plan.exists} \)}
                    \State \( \Pi.\text{UpdateWithPlan}(\text{plan}) \)
                    \State \( X_{start} \leftarrow \text{plan.endPosition} \)
                    \State $\mathcal{A}$.UpdateAutomatonState($\mathcal{S'}$)
                \Else
                    \State \( \Pi, \mathcal{A}, \text{path} \leftarrow \text{Backtrack}(\Pi, \mathcal{A}, \text{path}) \)
                \EndIf
            \EndIf
        \EndWhile
    \EndWhile
    \State \Return \( \Pi \)
\EndProcedure
\end{algorithmic}
\end{algorithm}

\begin{algorithm}[]
\caption{Task Progression Semantic Mapping Algorithm}
\label{alg:tpsm-algo}
\begin{algorithmic}[1]
\Procedure{GenerateTPSM}{$\mathcal{M}, \mathcal{M}_{rsm}, \mathcal{T}, \mathcal{A}, \mathcal{S'}, \theta$}
    \Statex \textbf{Input:}
    \Statex \quad \( \mathcal{M} \): Environment map.
    \Statex \quad \( \mathcal{M}_{rsm} \): Referent semantic map.
    \Statex \quad \( \mathcal{T} \): Automaton transition expression.
    \Statex \quad \( \mathcal{A} \): Task Automaton.
    \Statex \quad \( \mathcal{S'} \): Desired State.
    \Statex \quad \( \theta \): Nearness threshold.

    \Statex \textbf{Output:}
    \Statex \quad \( \mathcal{M}_{tpsm} \): Task Progression Semantic Map.

    \State \( \mathcal{M}_{tpsm} \leftarrow \text{Copy}(\mathcal{M}) \)
    \State \( \mathcal{P} \leftarrow \text{ExtractRelevantPredicates}(\mathcal{T}) \)

    \For{\( p \) in \( \mathcal{P} \)}
        \State \( \mathcal{R} \leftarrow \text{QueryPositions}(\mathcal{M}_{rsm}, p) \)
        \For{\( r \) in \( \mathcal{R} \)}
            \State \( G \leftarrow \left\{ g \mid g = r + \delta, \lVert \delta \rVert \leq \theta \right\} \)  \Comment{spherical grid of surrounding points}        
            \For{\( g \) in \( G \)}
                \State \( \mathcal{Q} \leftarrow \text{TruePredicatesAt}(g, \mathcal{M}_{rsm}, \theta) \)
                \State \( \mathcal{S}_{next} \leftarrow \text{ProgressAutomaton}(\mathcal{A}, \mathcal{Q}) \)
                \If{\( \mathcal{S}_{next} = \mathcal{S'} \)}
                    \State \( g.value \leftarrow 1 \) \Comment{Goal value}
                \ElsIf{\( \text{IsUndesired}(\mathcal{S}_{next}) \)}
                    \State \( g.value \leftarrow -1 \) \Comment{Avoidance value}
                \Else
                \State \( g.\text{value} \leftarrow 0 \)
                \EndIf
            \EndFor
            \State \( \text{AddPoints}(\mathcal{M}_{tpsm}, G) \)
        \EndFor
    \EndFor

    \State \Return \( \mathcal{M}_{tpsm} \)
\EndProcedure
\end{algorithmic}
\end{algorithm}

\section{Robot Skills}
\label{sec:robo-skills}
We define three predicate functions: \textbf{near}, \textbf{pick} and \textbf{release} for the navigation, picking and placing skills required for multi-object goal navigation and mobile pick-and-place. As highlighted in the main paper, we formalize navigation as continuous path planning problems and manipulation as object parameterized options. We discuss navigation at length in the paper, so here we focus on the pick and place manipulation skills.

\vspace{2mm}
\noindent\textbf{Pick Skill}: Once the robot has executed the near skill and is at the object to be manipulated, it takes a photo of the current environment to detect the object using the Owl-Vit model. The robot is guaranteed to be facing the object as the computed path plan uses the backprojected object 3D position to compute yaw angles for the robot. After detecting the object in the picture, we obtain a segmentation mask from detected boundary box using the Segment Anything model, and compute the center pixel of this mask. We feed this center pixel to the Boston dynamics grasping API to compute a motion plan to grasp the object. 

\vspace{2mm}
\noindent\textbf{Release Skill}: We implement a simple routine for the release skill which takes two parameters: the object to be placed and the place receptacle. Once a navigation skill gets the robot to the place receptacle, the robot gently moves its arm up or down to release the grasped object, based on the place receptable 3D position. Future work will implement more complex semantic placement strategies to better leverage LIMP's awareness and spatial grounding of instruction specific place receptacles. Kindly, visit our \href{https://robotlimp.github.io/}{\color{blue}{website}} to see demonstrations of these skills. \looseness=-1

\section{Evaluation and Baseline Details} 
\label{sec:eval-details}
All computation including planning, loading and running pretrained visual language models was done on a single computer equipped with one NVIDIA GeForce RTX 3090 GPU. We leverage GPT-4-0613 as the underlying LLM for our instruction understanding module and all our baselines. In all experiments we set the LLM temperature to 0, however since deterministic greedy token decoding is not guaranteed with GPT4, we perform three (3) queries for each instruction and evaluate on the most recurring response (mode response). \looseness=-1

We compare LIMP with baseline implementations of NLMap-Saycan \cite{chen_open-vocabulary_2023} and Code-as-policies \cite{liang_code_2023}. Both baselines use the same GPT-4 LLM, prompting structure, and in-context learning examples as our language understanding module. We integrate our composible referent descriptor syntax, spatial grounding module and low-level robot control into these baselines as APIs. This enables baselines to execute plans by querying relevant object positions, using our FMT* path planner to find paths to said positions and executing manipulation options. \looseness=-1

\begin{figure}
\centering
\includegraphics[scale=0.46]{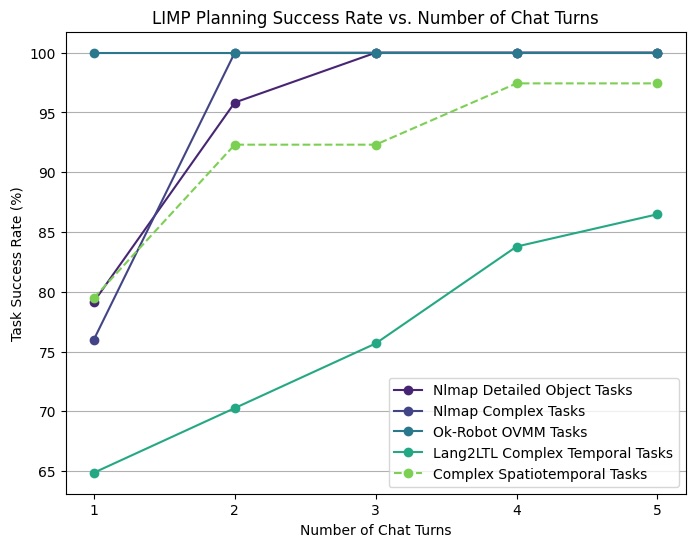}
\caption{Our interactive reprompting strategy implemented in the symbol verification node regenerates corrective formulas that improve plan success rates with minimal chat turns.}
\label{fig:language_module_chatturns}
\vspace*{-4mm}
\end{figure}

We visualize some qualitative results of LIMP from our experiments in Figure \ref{fig:qualitative}. We also highlight results in Figure \ref{fig:language_module_chatturns} that illustrates how our interactive symbol verification and reprompting strategy \ref{sec:interactive-verification-details} improves instruction satisfaction with minimal chat turns for different instruction sets. \looseness=-1

\begin{figure*}[t]
\centering
\vspace*{-1mm}
 \makebox[\textwidth][c]{\includegraphics[width=1.05\textwidth]{images/qual_result.jpeg}}%
   \vspace*{1mm}
  \caption{\textbf{[A]} Sample generated plan for a multi object-goal navigation task. \textbf{[B]} Sample generated plan for a mobile pick-and-place task.}
  \label{fig:qualitative}
    \vspace*{-0.3cm}
\end{figure*}

\subsection{NLMap-Saycan Implementation Prompt}
\begin{lstlisting}[language=gptprompt, caption= Exact prompt to implement NlMap-Saycan LLM planner]
You are an LLM for robot planning that understands logical operators such as &, |, ~ , etc. You have a finite set of robot predicates and spatial predicates, given a language instruction, your goal is to generate a sequence of actions that uses appropriate composition of robot and spatial predicates with relevant details from the instruction as arguments.
Robot predicate set (near,pick,release). 
Usage:
near[referent_1]:returns true if the desired spatial relationship is for robot to be near referent_1.
pick[referent_1]:can only execute picking skill on referent_1 and return True when near[referent_1].
release[referent_1,referent_2]:can only execute release skill on referent_1 and return True when near[referent_2].
Spatial predicate set (isbetween,isabove,isbelow,isleftof,isrightof,isnextto,isinfrontof,isbehind). 
Usage:
referent_1::isbetween(referent_2,referent_3):returns true if referent_1 is between referent_2 and referent_3.
referent_1::isabove(referent_2):returns True if referent_1 is above referent_2.
referent_1::isbelow(referent_2):returns True if referent_1 is below referent_2.
referent_1::isleftof(referent_2):returns True if referent_1 is left of referent_2.
referent_1::isrightof(referent_2):returns True if referent_1 is right of referent_2.
referent_1::isnextto(referent_2):returns True if referent_1 is close to referent_2.
referent_1::isinfrontof(referent_2):returns True if referent_1 is in front of referent_2.
referent_1::isbehind(referent_2):returns True if referent_1 is behind referent_2.
Rules:
Strictly only use the finite set of robot and spatial predicates!
Strictly stick to the usage format!
Compose spatial predicates where necessary!
You should strictly stick to mentioned objects, however you are allowed to propose and include plausible objects if and only if not mentioned in instruction but required based on context of instruction! 
Pay attention to instructions that require performing certain actions multiple times in generating and sequencing the predicates for the final Output!
Example:
Input_instruction: Go to the orange building but before that pass by the coffee shop, then go to the parking sign.
Output: 
1. near[coffee_shop] 
2. near[orange_building]
3. near[parking_sign]
Input_instruction: Go to the blue sofa then the laptop, after that bring me the brown bag between the television and the kettle on the left of the green seat, I am standing by the sink.
Output: 
1. near[blue_sofa]
2. near[laptop]
3. near[brown_bag::isbetween(television,kettle::isleftof(green_seat))] 
4. pick[brown_bag::isbetween(television,kettle::isleftof(green_seat))] 
5. near[sink] 
6. release[brown_bag,sink]
Input_instruction: Hey need you to pass by chair between the sofa and bag, pick up the bag and go to the orange napkin on the right of the sofa.
Output: 
1. near[chair::isbetween(sofa,bag)] 
2. near[bag] 
3. pick[bag]
4. near[orange_napkin::isrightof(sofa)]
Input_instruction:  Go to the chair between the green laptop and the yellow box underneath the play toy
Output: 
1. near[chair::isbetween(green_laptop,yellow_box::isbelow(play_toy))]
Input_instruction: Check the table behind the fridge and bring two beers to the couch one after the other
Output: 
1. near[table::isbehind(fridge)] 
2. pick[beer] 
3. near[couch] 
4. release[beer,couch] 
5. near[table::isbehind(fridge)] 
6. pick[beer] 
7. near[couch] 
8. release[beer,couch]
Input_instruction: <given_instruction>
Output:
\end{lstlisting}

\subsection{Code-as-Policies Implementation Prompt}
\begin{lstlisting}[language=gptprompt, caption= Exact prompt to implement Code-as-policies planner]
##Python robot planning script
from robotactions import near, pick, release
spatial_relationships = [
"isbetween", #referent_1::isbetween(referent_2,referent_3):returns true if referent_1 is between referent_2 and referent_3.
"isabove",   #referent_1::isabove(referent_2):returns True if referent_1 is above referent_2.
"isbelow",   #referent_1::isbelow(referent_2):returns True if referent_1 is below referent_2.
"isleftof",  #referent_1::isleftof(referent_2):returns True if referent_1 is left of referent_2.
"isrightof", #referent_1::isrightof(referent_2):returns True if referent_1 is right of referent_2.
"isnextto",  #referent_1::isnextto(referent_2):returns True if referent_1 is close to referent_2.
"isinfrontof", #referent_1::isinfrontof(referent_2):returns True if referent_1 is in front of referent_2.
"isbehind"  #referent_1::isbehind(referent_2):returns True if referent_1 is behind referent_2.]
##Rules:
##Strictly only use the finite set of robot and spatial predicates!
##Strictly stick to the usage format!
##Compose spatial predicates where necessary!
##You should strictly stick to mentioned objects, however you are allowed to propose and include plausible objects if and only if not mentioned in instruction but required based on context of instruction! 
##Pay attention to instructions that require performing certain actions multiple times in generating and sequencing the predicates for the final Output!
# Go to the orange building but before that pass by the coffee shop, then go to the parking sign.
ordered_navigation_goal_referents = ["coffee_shop", "orange_building", "parking_sign"]
for referent in ordered_navigation_goal_referents:
    near(referent)
# Go to the blue sofa then the laptop, after that bring me the brown bag between the television and the kettle on the left of the green seat, I am standing by the sink.
ordered_navigation_goal_referents = ["blue_sofa", "laptop", "brown_bag::isbetween(television,kettle::isleftof(green_seat))", "sink"]
referents_to_pick= ["brown_bag::isbetween(television,kettle::isleftof(green_seat))"]
release_location_referents = ["sink"]
picked_item = None
for referent in ordered_navigation_goal_referents:
    near(referent)
    if referent in referents_to_pick:
        pick(referent)
        picked_item = referent
    if referent in release_location_referents:
        release(picked_item, referent)
#Hey need you to pass by chair between the sofa and bag, pick up the bag and go to the orange napkin on the right of the sofa.
ordered_navigation_goal_referents = ["chair::isbetween(sofa,bag)", "bag", "orange_napkin::isrightof(sofa)"]
referents_to_pick= ["bag"]
picked_item = None
for referent in ordered_navigation_goal_referents:
    near(referent)
    if referent in referents_to_pick:
        pick(referent)
        picked_item = referent
#Go to the chair between the green laptop and the yellow box underneath the play toy
near("chair::isbetween(green_laptop,yellow_box::isbelow(play_toy))")
#Check the table behind the fridge and bring two beers to the couch one after the other
for i in range(2):
    near("table::isbehind(fridge)")
    pick("beer")
    near("couch")
    release("beer", "couch")

#<given_instruction>
\end{lstlisting}

\subsection{Instruction set}
\label{sec:instruction-set-details}
We perform a large-scale evaluation on 150 instructions across five real-world environments. The taskset includes 24 tasks with fine-grained object descriptions (NLMD) from \cite{chen_open-vocabulary_2023}, 25 tasks with complex language (NLMC) from \cite{chen_open-vocabulary_2023}, 25 tasks with simple structured phrasing (OKRB) from \cite{liu_demonstrating_2024}, 37 tasks with complex temporal structures (CT) from \cite{liu_grounding_2023}, and an additional 39 tasks we propose that have descriptive spatial constraints and temporal structures (CST).

\begin{lstlisting}[language=instructions, caption=Nlmap Detailed Object Tasks (NLMD)]
1: put the red can in the trash bin
2: put the brown multigrain chip bag in the woven basket
3: find the succulent plant
4: pick up the up side down mug
5: put the apple on the macbook with yellow stickers
6: use the dyson vacuum cleaner"
7: bring the kosher salt to the kitchen counter
8: put the used towels in washing machine
9: move the used mug to the dish washer
10: place the pickled cucumbers on the shelf
11: find my mug with the shape of a donut
12: put the almonds in the almond jar
13: fill the zisha tea pot with a coke from the cabinet
14: take the slippery floor sign with you
15: take the slippers that have holes on them to the shoe rack
16: find the mug on the mini fridge
17: bring the mint flavor gum to the small table
18: find some n95 masks
19: grab the banana with most black spots
20: fill the empty bottle with lemon juice
21: throw away the apple that's about to rot
22: throw away the rotting banana
23: take the box of organic blueberries out of the fridge
24: give a can of diet coke to the toy cat
\end{lstlisting}

\begin{lstlisting}[language=instructions, caption=Nlmap Complex Language Understanding Tasks (NLMC)]
1: I opened a pepsi earlier, bring an open can to the orange table
2: I spilled my coke, can you put a replacement on the kitchen counter
3: I spilled some coke on the television, go and bring something to clean it up
4: I accidentally dropped that jalapeno chips after eating it. Would you mind throwing it away
5: I like fruits, can you put something I would like on the yellow sofa for me
6: There is a green counter, a yellow counter, and a table. visit all the locations
7: There is a green counter, a trash can, and a table. visit all the locations
8: Redbull is my faviorite drink, can you put one on the desk please
9: Would you bring a coke can to the door for me
10: Please, move the pepsi to the red counter
11: Can you move the coke can to the orange counter
12: Would you throw away the bag of chips for me
13: Put an energy bar and water bottle on the table
14: Bring a lime soda and a bag of chips to the sofa
15: Can you throw away the apple and bring a coke to the bed
16: Bring a 7up can and a tea to the office desk
17: Move the multigrain chips to the table and an apple to the yellow counter
18: Move the lime soda, the sponge, and the water bottle to the table
19: Bring two sodas to the table
20: Move three cokes to the trash can
21: Throw away two cokes from the counter
22: Bring two different sodas to the cabinet, there is a coke, pepsi, soup, tea and 7up in the fridge
23: Bring an apple, a coke, and water bottle to the sofa
24: I spilled my coke on the table, throw it away and then bring something to help clean
25: I just worked out, can you bring me a drink and a snack to recover, i am on the sof
\end{lstlisting}

\begin{lstlisting}[language=instructions, caption=Ok-Robot Tasks (OKRB)]
1: Move the Takis on the desk to the nightstand
2: Move the soda can to the box
3: Move the purple shampoo to the red bag
4: Move the white meds box to the trash bin
5: Move the power adapter to the chair
6: Move the blue gloves to the sink
7: Move the McDonalds paper bag to the stove
8: Move the herbal tea can to the box
9: Move the cooking oil bottle to the marble surface
10: Move the milk bottle to the chair
11: Move the purple shampoo to the white rack
12: Move the purple lightbulb box to the sofa chair
13: Pick up purple medicine, drop it on chair
14: Pick up eyeglass case, drop it on chair
15: Pick up grey rag, drop it in sink
16: Pick up golden can rag, drop it on table
17: Pick up red navaratna oil, drop it on table
18: Pick up purple shampoo, drop it on green rack
19: Pick up taki chips, drop it on countertop
20: Pick up bandage box, drop it in dustbin
21: Pick up white aerosol, drop it in trash can
22: Pick up peanut butter, drop it on countertop
23: Pick up blue gloves, drop it in sink
24: Pick up brown box, drop it on chair
25: Pick up axe body spray, drop it on shel
\end{lstlisting}

\begin{lstlisting}[language=instructions, caption=Lang2LTL Complex temporal Tasks (CT)]
1: go to brown bookshelf, metal desk, wooden desk, kitchen counter, and the blue couch in any order
2: move to grey door, then bookrack, then go to the brown desk, then counter, then white desk  
3: visit brown wooden desk but only after bookshelf  
4: go from brown bookshelf to white metal desk and only visit each landmark one time  
5: go to brown wooden desk exactly once and do not visit brown desk before bookshelf  
6: go to brown desk only after visiting bookshelf, in addition go to brown desk only after visiting white desk  
7: visit the blue IKEA couch, in addition never go to the big steel door  
8: visit white kitchen counter then go to brown desk, in addition never visit white table  
9: go to the grey door, and only then go to the bookshelf, in addition always avoid the table  
10: go to kitchen counter then wooden desk, in addition after going to counter, you must avoid white table  
11: Go to bookshelf, alternatively go to metal desk  
12: Go to counter, alternatively go to metal desk  
13: Go to the counter, but never visit the counter  
14: do not go to the wooden desk until bookshelf, and do not go to bookshelf until wooden desk  
15: go to brown desk exactly once, in addition go to brown desk at least twice  
16: move to couch exactly twice, in addition pass by counter at most once
17: navigate to the counter then the brown desk, in addition after going to the counter, you must avoid doorway  
18: visit counter at least six times
19: either go to bookshelf then the brown desk, or go to couch
20: navigate to the wooden door, the glass door and the table, kitchen counter, and the blue couch in any order  
21: go to the painting, then find the kitchen table, then front desk, then staircase  
22: navigate to classroom but do not visit classroom before the white table  
23: only visit classroom once, and do not visit classroom until you visit elevator first  
24: Go to the staircase, front desk and the white table in that exact order. You are not permitted to revisit any of these locations  
25: go to the front desk then the yellow office door, in addition do not visit the glass door  
26: go to the stairs then the front desk, in addition avoid purple elevator  
27: move to elevator then front desk, in addition avoid staircase  
28: go to front desk then the cabinet, always avoid the elevator  
29: Go to elevator, alternatively go to staircase  
30: Visit the elevator exactly once, in addition visit the front desk on at least 2 separate occasions  
31: Go to the office, in addition avoid visiting the elevator and the classroom  
32: Visit the front desk, in addition you are not permitted to visit elevator and staircase  
33: Visit the purple door elevator, then go to the front desk and then go to the kitchen table, in addition you can never go to the elevator once you have seen the front desk
34: Visit the front desk then the sofa then the white table, in addition if you visit the sofa you must avoid the television after that
35: Go to the glass door, but never visit the glass door  
36: do not go to the white table until classroom, and do not go to the classroom until white table  
37: find the office, in addition avoid visiting the front desk and the classroom and the table
\end{lstlisting}

\begin{lstlisting}[language=instructions, caption=Complex Spatiotemporal Tasks (CST)]
1: Go to the red sofa but dont pass by any door then go to the computer, the sofa is behind the pillar and on the right side
2: Find the door with posters in front of the yellow pillar then go stand by the tree but avoid any sofa
3: There is a cabinet, a television and a tree in this room. I want you to go to the last item I mentioned then pass by the second item but dont go close to the first when you are going to thsecond item.",
4: Go to the trash bin, after that go to the large television. Actually can you go to the laptop on the table before doing all that?
5: Try to go to the tree without going near the trash bin.
6: Find the yellow trash bin and go to it, then go by the white table. But before doing the first thing visit where the fridge is then after the last thing, stop by the red sofa but avoid theooden door with posters",
7: Go to the television but avoid the tree.
8: Go to any trash bin but not the yellow one
9: Visit one of the televisions in this room then go to the fridge to the left of the cabinet
10: I need you to first go to the table between the pillar and the door with posters, then go to the tree. You know what, ignore the last thing I asked you to go to, after the first thing go tthe computer next to the green box instead.
11: Go to the red sofa then the blue one next to the cabinet but first pass by the book shelf
12: Go to water filter, dont pass by any computer to the left of the red sofa
13: Visit these things in the following order the microwave, then the black chair infront of the yellow robot then the fridge, but before doing any of this go to the robot
14: Go to the cabinet between the blue sofa and the brown box. 
15: I want you to go the cabinet but dont go near any black chair. Hold on, actually can you pass by the whiteboard before doing all that?
16: I need you to go to the red sofa but dont go anywhere near the sink
17: You can either go as close as you can to the sink or the plant pot but dont go near the yellow robot.
18: Go to the green curtain after that the large television but try not to go near the red sofa
19: Go to the sink then the red sofa, then the guitar. After that I want you to return to the first thing you visited.
20: I need you to go to the sink then any computer on a table but before you start that can you first go to the bookshelf, when you are done with all that, return to brown box
21: There are a couple of things in this room, a coffee machine, a robot, a computer and a plant pot. Visit the second item I mentioned then the first, after that visit the brown box, then go  the items you haven't visited yet.
22: Avoid any cabinet but go get the mug on the blue sofa and take it to red one
23: Please get me my plant pot from between the sofa and the sink, bring it to the white cabinet
24: Can you pick up my book from the left side of the television and bring it to me? I am sitting on the couch which is to the right of the door with posters. Make sure to never go near the trhbin while doing all this, it on the yellow table.
25: Bring the green plush toy to the whiteboard infront of it, watchout for the robot infront of the toy
26: Go get my yellow bag and bring it to the table between the yellow pillar and the wooden door with posters next to the whiteboard
27: Find the bottle that is on the table to the left of the computer and bring it to the wardrobe that is next to the glass door
28: Go to the cabinet between the blue sofa and the yellow robot or the cabinet to the left of the blue sofa. 
29: Go take the green toy that is next to the sofa under the poster and bring it to the bookshelf
30: I have a white cabinet, a green toy, a bookshelf and a red chair around here somewhere. Take the second item I mentioned from between the first item and the third. Bring it the cabinet butvoid the last item at all costs.
31: Hey I am standing by the whiteboard infront of the bookshelf, can you bring me the mug from the table to the left of the fridge?
32: Never pass by a robot, but i need you to bring the bag on the table to the cabinet
33: Take a soda can to the fridge, you can find one on the table to the left of the blue sofa
34: Hey, can you pick up my bag for me? It is under the table infront of the glass door, bring it to the cabinet. Actually you know what, forget what i asked for and bring the green toy instea its at the same place",
35: Go to the sink and take the mug beside the coffee maker, drop it off at the red sofa
36: Go to the sofa behind the chair next to the orange counter in front of the pillar
37: I want you to go visit the chair behind the sofa then the green bag, do that 3 times but during the second time avoid the fridge 
38: Go bring the plush toy between the sofa and the television to the couch, my cat is on the brown table so please dont pass near the table when you are returning with the toy
39: I think I left my wallet on the kitchen counter, go get it i will meet you at the bed. There is a lantern near the bed, make sure you dont hit it
\end{lstlisting}

%% file: root.bbl
\begin{thebibliography}{10}
\providecommand{\url}[1]{#1}
\csname url@rmstyle\endcsname
\providecommand{\newblock}{\relax}
\providecommand{\bibinfo}[2]{#2}
\providecommand\BIBentrySTDinterwordspacing{\spaceskip=0pt\relax}
\providecommand\BIBentryALTinterwordstretchfactor{4}
\providecommand\BIBentryALTinterwordspacing{\spaceskip=\fontdimen2\font plus
\BIBentryALTinterwordstretchfactor\fontdimen3\font minus \fontdimen4\font\relax}
\providecommand\BIBforeignlanguage[2]{{%
\expandafter\ifx\csname l@#1\endcsname\relax
\typeout{** WARNING: IEEEtran.bst: No hyphenation pattern has been}%
\typeout{** loaded for the language `#1'. Using the pattern for}%
\typeout{** the default language instead.}%
\else
\language=\csname l@#1\endcsname
\fi
#2}}

\bibitem{hu_toward_2024}
\BIBentryALTinterwordspacing
Y.~Hu, Q.~Xie, V.~Jain, J.~Francis, J.~Patrikar, N.~Keetha, S.~Kim, Y.~Xie, T.~Zhang, H.-S. Fang, S.~Zhao, S.~Omidshafiei, D.-K. Kim, A.-a. Agha-mohammadi, K.~Sycara, M.~Johnson-Roberson, D.~Batra, X.~Wang, S.~Scherer, C.~Wang, Z.~Kira, F.~Xia, and Y.~Bisk, ``Toward {General}-{Purpose} {Robots} via {Foundation} {Models}: {A} {Survey} and {Meta}-{Analysis},'' Oct. 2024, arXiv:2312.08782 [cs]. [Online]. Available: \url{http://arxiv.org/abs/2312.08782}
\BIBentrySTDinterwordspacing

\bibitem{firoozi_foundation_2024}
\BIBentryALTinterwordspacing
R.~Firoozi, J.~Tucker, S.~Tian, A.~Majumdar, J.~Sun, W.~Liu, Y.~Zhu, S.~Song, A.~Kapoor, K.~Hausman, B.~Ichter, D.~Driess, J.~Wu, C.~Lu, and M.~Schwager, ``\BIBforeignlanguage{en}{Foundation models in robotics: {Applications}, challenges, and the future},'' \emph{\BIBforeignlanguage{en}{The International Journal of Robotics Research}}, p. 02783649241281508, Sept. 2024, publisher: SAGE Publications Ltd STM. [Online]. Available: \url{https://doi.org/10.1177/02783649241281508}
\BIBentrySTDinterwordspacing

\bibitem{gadre_cows_2023}
\BIBentryALTinterwordspacing
S.~Y. Gadre, M.~Wortsman, G.~Ilharco, L.~Schmidt, and S.~Song, ``\BIBforeignlanguage{en}{{CoWs} on {Pasture}: {Baselines} and {Benchmarks} for {Language}-{Driven} {Zero}-{Shot} {Object} {Navigation}},'' in \emph{\BIBforeignlanguage{en}{2023 {IEEE}/{CVF} {Conference} on {Computer} {Vision} and {Pattern} {Recognition} ({CVPR})}}.\hskip 1em plus 0.5em minus 0.4em\relax IEEE, June 2023, pp. 23\,171--23\,181. [Online]. Available: \url{https://ieeexplore.ieee.org/document/10203853/}
\BIBentrySTDinterwordspacing

\bibitem{shah_lm-nav_2023}
\BIBentryALTinterwordspacing
D.~Shah, B.~Osiński, B.~Ichter, and S.~Levine, ``\BIBforeignlanguage{en}{{LM}-{Nav}: {Robotic} {Navigation} with {Large} {Pre}-{Trained} {Models} of {Language}, {Vision}, and {Action}},'' in \emph{\BIBforeignlanguage{en}{Proceedings of {The} 6th {Conference} on {Robot} {Learning}}}.\hskip 1em plus 0.5em minus 0.4em\relax PMLR, Mar. 2023, pp. 492--504, iSSN: 2640-3498. [Online]. Available: \url{https://proceedings.mlr.press/v205/shah23b.html}
\BIBentrySTDinterwordspacing

\bibitem{huang_visual_2023}
\BIBentryALTinterwordspacing
C.~Huang, O.~Mees, A.~Zeng, and W.~Burgard, ``Visual {Language} {Maps} for {Robot} {Navigation},'' in \emph{2023 {IEEE} {International} {Conference} on {Robotics} and {Automation} ({ICRA})}, May 2023, pp. 10\,608--10\,615. [Online]. Available: \url{https://ieeexplore.ieee.org/document/10160969}
\BIBentrySTDinterwordspacing

\bibitem{huang_audio_2024}
------, ``\BIBforeignlanguage{en}{Audio {Visual} {Language} {Maps} for {Robot} {Navigation}},'' in \emph{\BIBforeignlanguage{en}{Experimental {Robotics}}}, M.~H. Ang~Jr and O.~Khatib, Eds.\hskip 1em plus 0.5em minus 0.4em\relax Cham: Springer Nature Switzerland, 2024, pp. 105--117.

\bibitem{liu_grounding_2023}
\BIBentryALTinterwordspacing
J.~X. Liu, Z.~Yang, I.~Idrees, S.~Liang, B.~Schornstein, S.~Tellex, and A.~Shah, ``\BIBforeignlanguage{en}{Grounding {Complex} {Natural} {Language} {Commands} for {Temporal} {Tasks} in {Unseen} {Environments}},'' in \emph{\BIBforeignlanguage{en}{Proceedings of {The} 7th {Conference} on {Robot} {Learning}}}.\hskip 1em plus 0.5em minus 0.4em\relax PMLR, Dec. 2023, pp. 1084--1110, iSSN: 2640-3498. [Online]. Available: \url{https://proceedings.mlr.press/v229/liu23d.html}
\BIBentrySTDinterwordspacing

\bibitem{anderson_evaluation_2018}
\BIBentryALTinterwordspacing
P.~Anderson, A.~Chang, D.~S. Chaplot, A.~Dosovitskiy, S.~Gupta, V.~Koltun, J.~Kosecka, J.~Malik, R.~Mottaghi, M.~Savva, and A.~R. Zamir, ``On {Evaluation} of {Embodied} {Navigation} {Agents},'' July 2018, arXiv:1807.06757 [cs]. [Online]. Available: \url{http://arxiv.org/abs/1807.06757}
\BIBentrySTDinterwordspacing

\bibitem{yenamandra_homerobot_2023}
\BIBentryALTinterwordspacing
S.~Yenamandra, A.~Ramachandran, K.~Yadav, A.~S. Wang, M.~Khanna, T.~Gervet, T.-Y. Yang, V.~Jain, A.~Clegg, J.~M. Turner, Z.~Kira, M.~Savva, A.~X. Chang, D.~S. Chaplot, D.~Batra, R.~Mottaghi, Y.~Bisk, and C.~Paxton, ``\BIBforeignlanguage{en}{{HomeRobot}: {Open}-{Vocabulary} {Mobile} {Manipulation}},'' in \emph{\BIBforeignlanguage{en}{Proceedings of {The} 7th {Conference} on {Robot} {Learning}}}.\hskip 1em plus 0.5em minus 0.4em\relax PMLR, Dec. 2023, pp. 1975--2011, iSSN: 2640-3498. [Online]. Available: \url{https://proceedings.mlr.press/v229/yenamandra23a.html}
\BIBentrySTDinterwordspacing

\bibitem{chen_open-vocabulary_2023}
\BIBentryALTinterwordspacing
B.~Chen, F.~Xia, B.~Ichter, K.~Rao, K.~Gopalakrishnan, M.~S. Ryoo, A.~Stone, and D.~Kappler, ``Open-vocabulary {Queryable} {Scene} {Representations} for {Real} {World} {Planning},'' in \emph{2023 {IEEE} {International} {Conference} on {Robotics} and {Automation} ({ICRA})}, May 2023, pp. 11\,509--11\,522. [Online]. Available: \url{https://ieeexplore.ieee.org/document/10161534}
\BIBentrySTDinterwordspacing

\bibitem{liu_demonstrating_2024}
\BIBentryALTinterwordspacing
P.~Liu, Y.~Orru, J.~Vakil, C.~Paxton, N.~Shafiullah, and L.~Pinto, ``\BIBforeignlanguage{en}{Demonstrating {OK}-{Robot}: {What} {Really} {Matters} in {Integrating} {Open}-{Knowledge} {Models} for {Robotics}},'' in \emph{\BIBforeignlanguage{en}{Robotics: {Science} and {Systems}}}.\hskip 1em plus 0.5em minus 0.4em\relax Robotics: Science and Systems Foundation, July 2024. [Online]. Available: \url{http://www.roboticsproceedings.org/rss20/p091.pdf}
\BIBentrySTDinterwordspacing

\bibitem{liang_code_2023}
\BIBentryALTinterwordspacing
J.~Liang, W.~Huang, F.~Xia, P.~Xu, K.~Hausman, B.~Ichter, P.~Florence, and A.~Zeng, ``Code as {Policies}: {Language} {Model} {Programs} for {Embodied} {Control},'' in \emph{2023 {IEEE} {International} {Conference} on {Robotics} and {Automation} ({ICRA})}, May 2023, pp. 9493--9500. [Online]. Available: \url{https://ieeexplore.ieee.org/document/10160591}
\BIBentrySTDinterwordspacing

\bibitem{emerson_temporal_1990}
\BIBentryALTinterwordspacing
E.~A. Emerson, ``Temporal and {Modal} {Logic},'' in \emph{Handbook of {Theoretical} {Computer} {Science}, {Volume} {B}: {Formal} {Models} and {Semantics}}, J.~v. Leeuwen, Ed.\hskip 1em plus 0.5em minus 0.4em\relax Elsevier and MIT Press, 1990, pp. 995--1072. [Online]. Available: \url{https://doi.org/10.1016/b978-0-444-88074-1.50021-4}
\BIBentrySTDinterwordspacing

\bibitem{pan_data-efficient_2023}
\BIBentryALTinterwordspacing
J.~Pan, G.~Chou, and D.~Berenson, ``Data-{Efficient} {Learning} of {Natural} {Language} to {Linear} {Temporal} {Logic} {Translators} for {Robot} {Task} {Specification},'' in \emph{2023 {IEEE} {International} {Conference} on {Robotics} and {Automation} ({ICRA})}, May 2023, pp. 11\,554--11\,561. [Online]. Available: \url{https://ieeexplore.ieee.org/document/10161125}
\BIBentrySTDinterwordspacing

\bibitem{quartey_exploiting_2023}
\BIBentryALTinterwordspacing
B.~Quartey, A.~Shah, and G.~Konidaris, ``Exploiting {Contextual} {Structure} to {Generate} {Useful} {Auxiliary} {Tasks},'' in \emph{NeurIPS 2023 Workshop on Generalization in Planning}, vol. abs/2303.05038, 2023, arXiv: 2303.05038. [Online]. Available: \url{https://doi.org/10.48550/arXiv.2303.05038}
\BIBentrySTDinterwordspacing

\bibitem{liu_lang2ltl-2_2024}
\BIBentryALTinterwordspacing
J.~X. Liu, A.~Shah, G.~Konidaris, S.~Tellex, and D.~Paulius, ``{Lang2LTL}-2: {Grounding} {Spatiotemporal} {Navigation} {Commands} {Using} {Large} {Language} and {Vision}-{Language} {Models},'' in \emph{2024 {IEEE}/{RSJ} {International} {Conference} on {Intelligent} {Robots} and {Systems} ({IROS})}, Oct. 2024, pp. 2325--2332, iSSN: 2153-0866. [Online]. Available: \url{https://ieeexplore.ieee.org/document/10802696}
\BIBentrySTDinterwordspacing

\bibitem{rosen_synthesizing_2023}
\BIBentryALTinterwordspacing
E.~Rosen, S.~James, S.~Orozco, V.~Gupta, M.~Merlin, S.~Tellex, and G.~Konidaris, ``\BIBforeignlanguage{en}{Synthesizing {Navigation} {Abstractions} for {Planning} with {Portable} {Manipulation} {Skills}},'' in \emph{\BIBforeignlanguage{en}{Proceedings of {The} 7th {Conference} on {Robot} {Learning}}}.\hskip 1em plus 0.5em minus 0.4em\relax PMLR, Dec. 2023, pp. 2278--2287, iSSN: 2640-3498. [Online]. Available: \url{https://proceedings.mlr.press/v229/rosen23a.html}
\BIBentrySTDinterwordspacing

\bibitem{pnueli_temporal_1977}
\BIBentryALTinterwordspacing
A.~Pnueli, ``The temporal logic of programs,'' in \emph{18th {Annual} {Symposium} on {Foundations} of {Computer} {Science} (sfcs 1977)}, Oct. 1977, pp. 46--57, iSSN: 0272-5428. [Online]. Available: \url{https://ieeexplore.ieee.org/document/4567924}
\BIBentrySTDinterwordspacing

\bibitem{menghi_specification_2021}
\BIBentryALTinterwordspacing
C.~Menghi, C.~Tsigkanos, P.~Pelliccione, C.~Ghezzi, and T.~Berger, ``{ Specification Patterns for Robotic Missions },'' \emph{IEEE Transactions on Software Engineering}, vol.~47, no.~10, pp. 2208--2224, Oct. 2021. [Online]. Available: \url{https://doi.ieeecomputersociety.org/10.1109/TSE.2019.2945329}
\BIBentrySTDinterwordspacing

\bibitem{berg_grounding_2020}
\BIBentryALTinterwordspacing
M.~Berg, D.~Bayazit, R.~Mathew, A.~Rotter-Aboyoun, E.~Pavlick, and S.~Tellex, ``Grounding {Language} to {Landmarks} in {Arbitrary} {Outdoor} {Environments},'' in \emph{2020 {IEEE} {International} {Conference} on {Robotics} and {Automation} ({ICRA})}, May 2020, pp. 208--215, iSSN: 2577-087X. [Online]. Available: \url{https://ieeexplore.ieee.org/document/9197068}
\BIBentrySTDinterwordspacing

\bibitem{cosler_nl2spec_2023}
M.~Cosler, C.~Hahn, D.~Mendoza, F.~Schmitt, and C.~Trippel, ``\BIBforeignlanguage{en}{nl2spec: {Interactively} {Translating} {Unstructured} {Natural} {Language} to {Temporal} {Logics} with {Large} {Language} {Models}},'' in \emph{\BIBforeignlanguage{en}{Computer {Aided} {Verification}}}, C.~Enea and A.~Lal, Eds.\hskip 1em plus 0.5em minus 0.4em\relax Cham: Springer Nature Switzerland, 2023, pp. 383--396.

\bibitem{fuggitti_nl2ltl_2023}
\BIBentryALTinterwordspacing
F.~Fuggitti and T.~Chakraborti, ``\BIBforeignlanguage{en}{{NL2LTL} – a {Python} {Package} for {Converting} {Natural} {Language} ({NL}) {Instructions} to {Linear} {Temporal} {Logic} ({LTL}) {Formulas}},'' \emph{\BIBforeignlanguage{en}{Proceedings of the AAAI Conference on Artificial Intelligence}}, vol.~37, no.~13, pp. 16\,428--16\,430, 2023, number: 13. [Online]. Available: \url{https://ojs.aaai.org/index.php/AAAI/article/view/27068}
\BIBentrySTDinterwordspacing

\bibitem{chen_nl2tl_2023}
\BIBentryALTinterwordspacing
Y.~Chen, R.~Gandhi, Y.~Zhang, and C.~Fan, ``{NL2TL}: {Transforming} {Natural} {Languages} to {Temporal} {Logics} using {Large} {Language} {Models},'' in \emph{Proceedings of the 2023 {Conference} on {Empirical} {Methods} in {Natural} {Language} {Processing}}, H.~Bouamor, J.~Pino, and K.~Bali, Eds.\hskip 1em plus 0.5em minus 0.4em\relax Singapore: Association for Computational Linguistics, Dec. 2023, pp. 15\,880--15\,903. [Online]. Available: \url{https://aclanthology.org/2023.emnlp-main.985/}
\BIBentrySTDinterwordspacing

\bibitem{vardi_automata-theoretic_1996}
\BIBentryALTinterwordspacing
M.~Y. Vardi, ``\BIBforeignlanguage{en}{An automata-theoretic approach to linear temporal logic},'' in \emph{\BIBforeignlanguage{en}{Logics for {Concurrency}: {Structure} versus {Automata}}}, F.~Moller and G.~Birtwistle, Eds.\hskip 1em plus 0.5em minus 0.4em\relax Berlin, Heidelberg: Springer, 1996, pp. 238--266. [Online]. Available: \url{https://doi.org/10.1007/3-540-60915-6_6}
\BIBentrySTDinterwordspacing

\bibitem{kress-gazit_temporal-logic-based_2009}
\BIBentryALTinterwordspacing
H.~Kress-Gazit, G.~E. Fainekos, and G.~J. Pappas, ``Temporal-{Logic}-{Based} {Reactive} {Mission} and {Motion} {Planning},'' \emph{IEEE Transactions on Robotics}, vol.~25, no.~6, pp. 1370--1381, Dec. 2009, conference Name: IEEE Transactions on Robotics. [Online]. Available: \url{https://ieeexplore.ieee.org/document/5238617}
\BIBentrySTDinterwordspacing

\bibitem{colledanchise_synthesis_2017}
\BIBentryALTinterwordspacing
M.~Colledanchise, R.~M. Murray, and P.~Ögren, ``Synthesis of correct-by-construction behavior trees,'' in \emph{2017 {IEEE}/{RSJ} {International} {Conference} on {Intelligent} {Robots} and {Systems} ({IROS})}, Sept. 2017, pp. 6039--6046, iSSN: 2153-0866. [Online]. Available: \url{https://ieeexplore.ieee.org/document/8206502}
\BIBentrySTDinterwordspacing

\bibitem{sutton_between_1999}
\BIBentryALTinterwordspacing
R.~S. Sutton, D.~Precup, and S.~Singh, ``Between {MDPs} and semi-{MDPs}: {A} framework for temporal abstraction in reinforcement learning,'' \emph{Artificial Intelligence}, vol. 112, no.~1, pp. 181--211, Aug. 1999. [Online]. Available: \url{https://www.sciencedirect.com/science/article/pii/S0004370299000521}
\BIBentrySTDinterwordspacing

\bibitem{yokoyama_asc_2024}
\BIBentryALTinterwordspacing
N.~Yokoyama, A.~Clegg, J.~Truong, E.~Undersander, T.-Y. Yang, S.~Arnaud, S.~Ha, D.~Batra, and A.~Rai, ``{ASC}: {Adaptive} {Skill} {Coordination} for {Robotic} {Mobile} {Manipulation},'' \emph{IEEE Robotics and Automation Letters}, vol.~9, no.~1, pp. 779--786, Jan. 2024, conference Name: IEEE Robotics and Automation Letters. [Online]. Available: \url{https://ieeexplore.ieee.org/document/10328058}
\BIBentrySTDinterwordspacing

\bibitem{jatavallabhula_conceptfusion_2023}
\BIBentryALTinterwordspacing
K.~Jatavallabhula, A.~Kuwajerwala, Q.~Gu, M.~Omama, G.~Iyer, S.~Saryazdi, T.~Chen, A.~Maalouf, S.~Li, N.~Keetha, A.~Tewari, J.~Tenenbaum, C.~Melo, M.~Krishna, L.~Paull, F.~Shkurti, and A.~Torralba, ``\BIBforeignlanguage{en}{{ConceptFusion}: {Open}-set multimodal {3D} mapping},'' in \emph{\BIBforeignlanguage{en}{Robotics: {Science} and {Systems} {XIX}}}.\hskip 1em plus 0.5em minus 0.4em\relax Robotics: Science and Systems Foundation, July 2023. [Online]. Available: \url{http://www.roboticsproceedings.org/rss19/p066.pdf}
\BIBentrySTDinterwordspacing

\bibitem{yang_plug_2024}
\BIBentryALTinterwordspacing
Z.~Yang, S.~S. Raman, A.~Shah, and S.~Tellex, ``Plug in the {Safety} {Chip}: {Enforcing} {Constraints} for {LLM}-driven {Robot} {Agents},'' in \emph{2024 {IEEE} {International} {Conference} on {Robotics} and {Automation} ({ICRA})}, May 2024, pp. 14\,435--14\,442. [Online]. Available: \url{https://ieeexplore.ieee.org/document/10611447}
\BIBentrySTDinterwordspacing

\bibitem{international2000functional}
I.~E. Commission \emph{et~al.}, ``Functional safety of electrical/electronic/programmable electronic safety related systems,'' \emph{IEC 61508}, 2000.

\bibitem{minderer_simple_2022}
M.~Minderer, A.~Gritsenko, A.~Stone, M.~Neumann, D.~Weissenborn, A.~Dosovitskiy, A.~Mahendran, A.~Arnab, M.~Dehghani, Z.~Shen, X.~Wang, X.~Zhai, T.~Kipf, and N.~Houlsby, ``\BIBforeignlanguage{en}{Simple {Open}-{Vocabulary} {Object} {Detection}},'' in \emph{\BIBforeignlanguage{en}{Computer {Vision} – {ECCV} 2022}}, S.~Avidan, G.~Brostow, M.~Cissé, G.~M. Farinella, and T.~Hassner, Eds.\hskip 1em plus 0.5em minus 0.4em\relax Cham: Springer Nature Switzerland, 2022, pp. 728--755.

\bibitem{kirillov_segment_2023}
\BIBentryALTinterwordspacing
A.~Kirillov, E.~Mintun, N.~Ravi, H.~Mao, C.~Rolland, L.~Gustafson, T.~Xiao, S.~Whitehead, A.~C. Berg, W.-Y. Lo, P.~Dollár, and R.~Girshick, ``Segment {Anything},'' in \emph{2023 {IEEE}/{CVF} {International} {Conference} on {Computer} {Vision} ({ICCV})}, Oct. 2023, pp. 3992--4003, iSSN: 2380-7504. [Online]. Available: \url{https://ieeexplore.ieee.org/document/10378323}
\BIBentrySTDinterwordspacing

\bibitem{majid_can_2004}
\BIBentryALTinterwordspacing
A.~Majid, M.~Bowerman, S.~Kita, D.~B.~M. Haun, and S.~C. Levinson, ``Can language restructure cognition? {The} case for space,'' \emph{Trends in Cognitive Sciences}, vol.~8, no.~3, pp. 108--114, Mar. 2004. [Online]. Available: \url{https://www.sciencedirect.com/science/article/pii/S1364661304000208}
\BIBentrySTDinterwordspacing

\bibitem{janson_fast_2015}
\BIBentryALTinterwordspacing
L.~Janson, E.~Schmerling, A.~Clark, and M.~Pavone, ``\BIBforeignlanguage{en}{Fast marching tree: {A} fast marching sampling-based method for optimal motion planning in many dimensions},'' \emph{\BIBforeignlanguage{en}{The International Journal of Robotics Research}}, vol.~34, no.~7, pp. 883--921, June 2015, publisher: SAGE Publications Ltd STM. [Online]. Available: \url{https://doi.org/10.1177/0278364915577958}
\BIBentrySTDinterwordspacing

\bibitem{tellex_robots_2020}
S.~Tellex, N.~Gopalan, H.~Kress-Gazit, and C.~Matuszek, ``Robots that use language,'' \emph{Annual Review of Control, Robotics, and Autonomous Systems}, vol.~3, pp. 25--55, 2020.

\bibitem{blukis_few-shot_2021}
\BIBentryALTinterwordspacing
V.~Blukis, R.~Knepper, and Y.~Artzi, ``\BIBforeignlanguage{en}{Few-shot {Object} {Grounding} and {Mapping} for {Natural} {Language} {Robot} {Instruction} {Following}},'' in \emph{\BIBforeignlanguage{en}{Proceedings of the 2020 {Conference} on {Robot} {Learning}}}.\hskip 1em plus 0.5em minus 0.4em\relax PMLR, Oct. 2021, pp. 1829--1854, iSSN: 2640-3498. [Online]. Available: \url{https://proceedings.mlr.press/v155/blukis21a.html}
\BIBentrySTDinterwordspacing

\bibitem{patel_grounding_2020}
\BIBentryALTinterwordspacing
R.~Patel, E.~Pavlick, and S.~Tellex, ``\BIBforeignlanguage{en}{Grounding {Language} to {Non}-{Markovian} {Tasks} with {No} {Supervision} of {Task} {Specifications}},'' in \emph{\BIBforeignlanguage{en}{Robotics: {Science} and {Systems} {XVI}}}.\hskip 1em plus 0.5em minus 0.4em\relax Robotics: Science and Systems Foundation, July 2020. [Online]. Available: \url{http://www.roboticsproceedings.org/rss16/p016.pdf}
\BIBentrySTDinterwordspacing

\bibitem{wang_learning_2021}
\BIBentryALTinterwordspacing
C.~Wang, C.~Ross, Y.-L. Kuo, B.~Katz, and A.~Barbu, ``\BIBforeignlanguage{en}{Learning a natural-language to {LTL} executable semantic parser for grounded robotics},'' in \emph{\BIBforeignlanguage{en}{Proceedings of the 2020 {Conference} on {Robot} {Learning}}}.\hskip 1em plus 0.5em minus 0.4em\relax PMLR, Oct. 2021, pp. 1706--1718, iSSN: 2640-3498. [Online]. Available: \url{https://proceedings.mlr.press/v155/wang21g.html}
\BIBentrySTDinterwordspacing

\bibitem{zheng_spatial_2021}
\BIBentryALTinterwordspacing
K.~Zheng, D.~Bayazit, R.~Mathew, E.~Pavlick, and S.~Tellex, ``Spatial {Language} {Understanding} for {Object} {Search} in {Partially} {Observed} {City}-scale {Environments},'' \emph{2021 30th IEEE International Conference on Robot \& Human Interactive Communication (RO-MAN)}, pp. 315--322, Aug. 2021, conference Name: 2021 30th IEEE International Conference on Robot \& Human Interactive Communication (RO-MAN) ISBN: 9781665404921 Place: Vancouver, BC, Canada Publisher: IEEE. [Online]. Available: \url{https://ieeexplore.ieee.org/document/9515426/}
\BIBentrySTDinterwordspacing

\bibitem{wang_lana_2023}
\BIBentryALTinterwordspacing
X.~Wang, W.~Wang, J.~Shao, and Y.~Yang, ``\BIBforeignlanguage{en}{{LANA}: {A} {Language}-{Capable} {Navigator} for {Instruction} {Following} and {Generation}},'' in \emph{\BIBforeignlanguage{en}{2023 {IEEE}/{CVF} {Conference} on {Computer} {Vision} and {Pattern} {Recognition} ({CVPR})}}.\hskip 1em plus 0.5em minus 0.4em\relax Vancouver, BC, Canada: IEEE, June 2023, pp. 19\,048--19\,058. [Online]. Available: \url{https://ieeexplore.ieee.org/document/10203301/}
\BIBentrySTDinterwordspacing

\bibitem{park_visual_2023}
\BIBentryALTinterwordspacing
S.-M. Park and Y.-G. Kim, ``\BIBforeignlanguage{en}{Visual language navigation: a survey and open challenges},'' \emph{\BIBforeignlanguage{en}{Artificial Intelligence Review}}, vol.~56, no.~1, pp. 365--427, Jan. 2023. [Online]. Available: \url{https://doi.org/10.1007/s10462-022-10174-9}
\BIBentrySTDinterwordspacing

\bibitem{yu_l3mvn_2023}
\BIBentryALTinterwordspacing
B.~Yu, H.~Kasaei, and M.~Cao, ``{L3MVN}: {Leveraging} {Large} {Language} {Models} for {Visual} {Target} {Navigation},'' in \emph{2023 {IEEE}/{RSJ} {International} {Conference} on {Intelligent} {Robots} and {Systems} ({IROS})}, Oct. 2023, pp. 3554--3560, iSSN: 2153-0866. [Online]. Available: \url{https://ieeexplore.ieee.org/document/10342512}
\BIBentrySTDinterwordspacing

\bibitem{song_llm-planner_2023}
\BIBentryALTinterwordspacing
C.~H. Song, B.~M. Sadler, J.~Wu, W.-L. Chao, C.~Washington, and Y.~Su, ``\BIBforeignlanguage{en}{{LLM}-{Planner}: {Few}-{Shot} {Grounded} {Planning} for {Embodied} {Agents} with {Large} {Language} {Models}},'' in \emph{\BIBforeignlanguage{en}{2023 {IEEE}/{CVF} {International} {Conference} on {Computer} {Vision} ({ICCV})}}.\hskip 1em plus 0.5em minus 0.4em\relax Paris, France: IEEE, Oct. 2023, pp. 2986--2997. [Online]. Available: \url{https://ieeexplore.ieee.org/document/10378628/}
\BIBentrySTDinterwordspacing

\bibitem{hsiung_generalizing_2022}
\BIBentryALTinterwordspacing
E.~Hsiung, H.~Mehta, J.~Chu, X.~Liu, R.~Patel, S.~Tellex, and G.~Konidaris, ``Generalizing to {New} {Domains} by {Mapping} {Natural} {Language} to {Lifted} {LTL},'' in \emph{2022 {International} {Conference} on {Robotics} and {Automation} ({ICRA})}.\hskip 1em plus 0.5em minus 0.4em\relax Philadelphia, PA, USA: IEEE Press, May 2022, pp. 3624--3630. [Online]. Available: \url{https://doi.org/10.1109/ICRA46639.2022.9812169}
\BIBentrySTDinterwordspacing

\bibitem{huang_voxposer_2023}
\BIBentryALTinterwordspacing
W.~Huang, C.~Wang, R.~Zhang, Y.~Li, J.~Wu, and L.~Fei-Fei, ``\BIBforeignlanguage{en}{{VoxPoser}: {Composable} {3D} {Value} {Maps} for {Robotic} {Manipulation} with {Language} {Models}},'' in \emph{\BIBforeignlanguage{en}{Proceedings of {The} 7th {Conference} on {Robot} {Learning}}}.\hskip 1em plus 0.5em minus 0.4em\relax PMLR, Dec. 2023, pp. 540--562, iSSN: 2640-3498. [Online]. Available: \url{https://proceedings.mlr.press/v229/huang23b.html}
\BIBentrySTDinterwordspacing

\bibitem{rana_sayplan_2023}
\BIBentryALTinterwordspacing
K.~Rana, J.~Haviland, S.~Garg, J.~Abou-Chakra, I.~Reid, and N.~Suenderhauf, ``\BIBforeignlanguage{en}{{SayPlan}: {Grounding} {Large} {Language} {Models} using {3D} {Scene} {Graphs} for {Scalable} {Robot} {Task} {Planning}},'' in \emph{\BIBforeignlanguage{en}{Proceedings of {The} 7th {Conference} on {Robot} {Learning}}}.\hskip 1em plus 0.5em minus 0.4em\relax PMLR, Dec. 2023, pp. 23--72, iSSN: 2640-3498. [Online]. Available: \url{https://proceedings.mlr.press/v229/rana23a.html}
\BIBentrySTDinterwordspacing

\bibitem{qiu_learning_2024}
\BIBentryALTinterwordspacing
R.-Z. Qiu, Y.~Hu, G.~Yang, Y.~Song, Y.~Fu, J.~Ye, J.~Mu, R.~Yang, N.~Atanasov, S.~A. Scherer, and X.~Wang, ``Learning {Generalizable} {Feature} {Fields} for {Mobile} {Manipulation},'' \emph{CoRR}, vol. abs/2403.07563, 2024, arXiv: 2403.07563. [Online]. Available: \url{https://doi.org/10.48550/arXiv.2403.07563}
\BIBentrySTDinterwordspacing

\bibitem{kostavelis_semantic_2015}
\BIBentryALTinterwordspacing
I.~Kostavelis and A.~Gasteratos, ``\BIBforeignlanguage{en}{Semantic mapping for mobile robotics tasks: {A} survey},'' \emph{\BIBforeignlanguage{en}{Robotics and Autonomous Systems}}, vol.~66, pp. 86--103, Apr. 2015. [Online]. Available: \url{https://linkinghub.elsevier.com/retrieve/pii/S0921889014003030}
\BIBentrySTDinterwordspacing

\bibitem{crespo_semantic_2020}
\BIBentryALTinterwordspacing
J.~Crespo, J.~C. Castillo, O.~M. Mozos, and R.~Barber, ``\BIBforeignlanguage{en}{Semantic {Information} for {Robot} {Navigation}: {A} {Survey}},'' \emph{\BIBforeignlanguage{en}{Applied Sciences}}, vol.~10, no.~2, p. 497, Jan. 2020, number: 2 Publisher: Multidisciplinary Digital Publishing Institute. [Online]. Available: \url{https://www.mdpi.com/2076-3417/10/2/497}
\BIBentrySTDinterwordspacing

\bibitem{pronobis_semantic_2011}
A.~Pronobis, P.~Jensfelt, and J.~Little, \emph{Semantic {Mapping} with {Mobile} {Robots}}.\hskip 1em plus 0.5em minus 0.4em\relax Stockholm: KTH Royal Institute of Technology, 2011.

\bibitem{fikes_strips_1971}
R.~E. Fikes and N.~J. Nilsson, ``Strips: A new approach to the application of theorem proving to problem solving,'' \emph{Artificial intelligence}, vol.~2, no. 3-4, pp. 189--208, 1971.

\bibitem{garrett_integrated_2021}
\BIBentryALTinterwordspacing
C.~R. Garrett, R.~Chitnis, R.~Holladay, B.~Kim, T.~Silver, L.~P. Kaelbling, and T.~Lozano-Pérez, ``\BIBforeignlanguage{en-US}{Integrated {Task} and {Motion} {Planning}},'' \emph{\BIBforeignlanguage{en-US}{Annual Review of Control, Robotics, and Autonomous Systems}}, vol.~4, no. Volume 4, 2021, pp. 265--293, May 2021, publisher: Annual Reviews. [Online]. Available: \url{https://www.annualreviews.org/content/journals/10.1146/annurev-control-091420-084139}
\BIBentrySTDinterwordspacing

\bibitem{garrett_pddlstream_2020}
\BIBentryALTinterwordspacing
C.~R. Garrett, T.~Lozano-Pérez, and L.~P. Kaelbling, ``\BIBforeignlanguage{en}{{PDDLStream}: {Integrating} {Symbolic} {Planners} and {Blackbox} {Samplers} via {Optimistic} {Adaptive} {Planning}},'' \emph{\BIBforeignlanguage{en}{Proceedings of the International Conference on Automated Planning and Scheduling}}, vol.~30, pp. 440--448, June 2020. [Online]. Available: \url{https://ojs.aaai.org/index.php/ICAPS/article/view/6739}
\BIBentrySTDinterwordspacing

\bibitem{holladay_planning_2021}
\BIBentryALTinterwordspacing
R.~Holladay, T.~Lozano-Pérez, and A.~Rodriguez, ``Planning for {Multi}-stage {Forceful} {Manipulation},'' in \emph{2021 {IEEE} {International} {Conference} on {Robotics} and {Automation} ({ICRA})}, May 2021, pp. 6556--6562, iSSN: 2577-087X. [Online]. Available: \url{https://ieeexplore.ieee.org/document/9561233}
\BIBentrySTDinterwordspacing

\end{thebibliography}
